\documentclass{article}
\usepackage{authblk}
\usepackage[utf8]{inputenc} 
\usepackage[T1]{fontenc}    
\usepackage{hyperref}       
\usepackage{url}            
\usepackage{booktabs}       
\usepackage{amsfonts}       
\usepackage{nicefrac}       
\usepackage{microtype}      
\usepackage{lipsum}
\usepackage{fancyhdr,graphicx,amsmath,amssymb}
\usepackage{hyperref}
\usepackage[usenames, dvipsnames]{color} 
\usepackage{enumerate, enumitem, amsthm, algorithm, algpseudocode, pifont, subfig, fullpage, csquotes, dashrule, tikz, bbm, booktabs, bm, verbatim}
\usepackage[framemethod=TikZ]{mdframed}
\usepackage[numbers]{natbib}

\newtheorem{defi}{Definition}
\newtheorem{coro}{Corollary}
\newtheorem{eg}{Example}
\newtheorem{rul}{Rule}

\newtheorem{theorem}{Theorem}

\newcommand{\indep}{\rotatebox[origin=c]{90}{$\models$}}

\definecolor{dblue}{RGB}{140, 212, 251}
\definecolor{dlblue}{RGB}{79, 142, 230}
\definecolor{dgray}{RGB}{205, 204, 204}
\definecolor{dred}{RGB}{239, 133, 125}
\definecolor{dorange}{RGB}{230, 169, 132}

\newcounter{WenhaoDefCounterTodo}
\newcounter{WenhaoDefCounterQuestion}
\newcounter{WenhaoDefCounterIdea}
\newcommand{\durl}[1]{\textcolor{dblue}{\underline{\url{#1}}}}

\newcommand{\ra}{\rightarrow}
\newcommand{\la}{\leftarrow}

\newmdenv[
  topline=false,
  bottomline=false,
  rightline = false,
  leftmargin=10pt,
  rightmargin=0pt,
  innertopmargin=0pt,
  innerbottommargin=0pt
]{innerproof}

\newcounter{DaveDefCounter}
\title{Causal Inference in medicine and in health policy, a summary}

\author{Wenhao Zhang}
\author{Ramin Ramezani}
\author{Arash Naeim}
\affil{Center for Smart Health, University of California, Los Angeles}

\begin{document}

\date{}
\maketitle


\label{chap:abstract}

\begin{abstract}
A data science task can be deemed as making sense of the data or testing a hypothesis about it. The conclusions inferred from data can greatly guide us to make informative decisions. Big data has enabled us to carry out countless prediction tasks in conjunction with machine learning, such as identifying high risk patients suffering from a certain disease and taking preventable measures. However, healthcare practitioners are not content with mere predictions – they are also interested in the cause-effect relation between input features and clinical outcomes. Understanding such relations will help doctors treat patients and reduce the risk effectively. Causality is typically identified by randomized controlled trials. Often such trials are not feasible when scientists and researchers turn to observational studies and attempt to draw inferences. However, observational studies may also be affected by selection and/or confounding biases that can result in wrong causal conclusions. In this chapter, we will try to highlight some of the drawbacks that may arise in traditional machine learning and statistical approaches to analyze the observational data, particularly in the healthcare data analytics domain. We will discuss causal inference and ways to discover the cause-effect from observational studies in healthcare domain. Moreover, we will demonstrate the applications of causal inference in tackling some common machine learning issues such as missing data and model transportability. Finally, we will discuss the possibility of integrating reinforcement learning with causality as a way to counter confounding bias.


\end{abstract}

\section{What is causality and why it matters}
\label{chap:chapter_intro}

In this section, we introduce the concept of causality and why causal reasoning is of paramount importance in analyzing data that arise in various domains such as healthcare or social sciences. 

\subsection{What is causality?}

The truism of :correlation does not imply causation" is well known and generally acknowledged \cite{geer2011correlation, havens1999correlation}. The question is how to define "causality".

\subsubsection{David Hume's definition}
The eighteenth-century philosopher, David Hume, defined causation in the language of the counterfactual: A is a cause of B, if: 
\begin{enumerate}
    \item B is always observed to follow A, and 
    \item A had not been, B never had existed \cite{hume2016enquiry}.
\end{enumerate}

In the former case, A is a sufficient causation of B, and A is a necessary causation of B in the latter case \cite{pearl2018book}. When both conditions are satisfied, we can safely say that A causes B (necessary-and-sufficient causation). For example, sunrise causes rooster’ crow. This cause-effect relation cannot be described the other way around. If the rooster is sick, the sunrise still occurs. Rooster’s crow is not a necessary causation of sunrise. Hence, rooster’s crow is an effect rather the cause of sunrise.

\subsubsection{Causality in medical research}
In medical research, the logical description of causality is considered too rigorous and occasionally not applicable. For example, smoking does not always lead to lung cancer. Causality in medical literature is often expressed in probabilistic terms \cite{morabia2005epidemiological,parascandola2011causes}. A type of chemotherapy treatment might increase the likelihood of survival of a patient diagnosed with cancer, but does not guarantee it.Therefore, we express our beliefs about the uncertainty about the real world in the language of probability. 

One main reason of probabilistic thinking is that we can easily quantify our beliefs in numeric values and build probabilistic models to explain the cause given our observation.  In clinical diagnosis, the doctors often seek the most plausible hypothesis (disease) that explain the evidence (symptom). Assume that a doctor observes a certain symptom S, and he or she has two explanations for this symptom, disease A or disease B.  If this doctor can quantify his or her belief into conditional probabilities, i.e., $Prob(disease\ A|Symptom\ S)$ and $Prob(disease\ B|Symptom\ S)$  (the likelihood of disease A or B may occur given the symptom S is observed). Then the doctor can choose the explanation that has larger value of conditional probability.

\paragraph{The dilemma: potential outcome framework}

When we study the causal-effect of a new treatment, we are interested in how the disease responses when we \textit{intervene} upon it. For example, a patient is likely to recover from the cancer when receiving a new type of chemotherapy treatment. To measure the causal effect on this particular patient, we shall compare the outcome of treatment to the outcome of no treatment. However, it is not possible to observe the two \textit{potential outcomes} of the same patient at once. Because this comparison is done using two parallel universes that we imagine: 1) a universe where the patient is treated with the new chemotherapy, and 2) the other where she is not treated. There is always one universe missing. This dilemma is known as the "fundamental problem of causal inference".

\subsection{Why causal inference?}

\begin{figure}[ht]
    \centering
    \includegraphics[width=70mm]{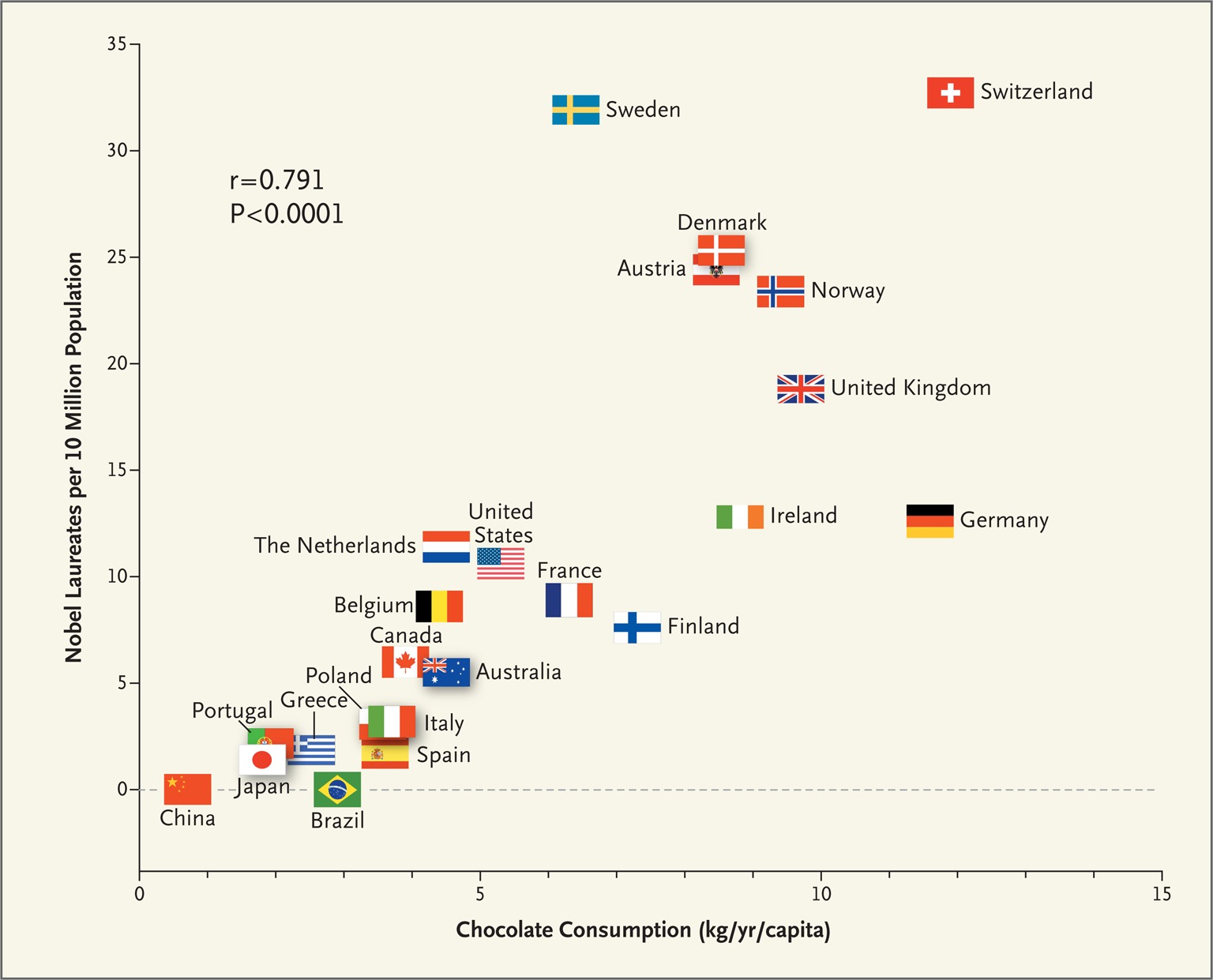}
    \caption{A spurious correlation between the chocolate consumption and the number of Nobel Laureates by countries \cite{chocolate}.}
    \label{fig:chocolate}
\end{figure}

A data science task can be deemed as making sense of the data or to test a hypothesis about it. The conclusions inferred from data can greatly guide us to make informative decisions. Big data has enabled us to carry out countless prediction tasks in conjunction with machine learning. However, there exist a large gap between highly accurate predictions and decision making. For example, an interesting study  \cite{chocolate} reports that there is a "surprisingly powerful correlation"  ($ \rho=0.79,\ p < 0.0001$) between the chocolate consumption and the number of Nobel Laureates in a country  (Fig. \ref{fig:chocolate}). The policy makers might hesitate to promote chocolate consumption as a way of obtaining more Nobel prizes. The developed western countries where people eat more chocolate are more likely to have better education systems and chocolate consumption has no direct impact on the number of Nobel Laureates. As such, intervening on chocolate cannot possibly lead us to desired outcome. In section \ref{chap:chapter_deconfound}, we will explore more examples of the spurious correlations explained by confounders (In statistics, confounder is a variable that impacts a dependent variable as well as an independent variable at the same time, causing a spurious correlation \cite{vanderweele2013defi}) and how to use causal inference to gauge the real causal effect between variables under such circumstances.

{}

In predictive tasks, understanding of causality, mechanisms through which an outcome is produced, will yield more descriptive models with better performance. In machine learning, for instance, one underlying assumption, generally, is that training and testing datasets have identical or at least comparable characteristics/distributions. This assumption is often violated in real practice. For example, an activity recognition model built on a training cohort of highly active participants might perform poorly if it is applied over a cohort of bedridden elderly patients. In this example, variables \textit{age} and \textit{mobility} are the causes that explain the difference between two datasets. Therefore, understanding causality between various features and outcomes is an integral part of a robust and generalized machine learning model. Often, most statistical models rely upon pure correlations perform well under static conditions where the characteristics of the dataset are invariant. Once the context changes, such correlations may no longer exist. On the contrary, relying on the causal relations between variables can produce models less prone to change with context. In section \ref{chap:chapter_transportability}, we will discuss on the external validity and transportability of machine learning models.

In many real-world data analytics, in addition to relying solely on statistical relations amongst data elements, it is essential for machine learning practitioners to ask questions surrounding "causal intervention" and "counterfactual reasoning". Questions such as  "what would Y be if I do X ?" (causal intervention) or "would the outcome change if I had acted differently?" (counterfactual).  Suppose that one wants to find the effects of wine consumption on heart disease. We certainly live in a world in which we cannot run randomized controlled trials asking people to drink wine, say, 1 glass every night, and force them to comply with it for a decade to find the effect of wine on heart disease. In such scenarios we normally resort to observational studies that may eventually highlight associations between wine consumption and reduced risk of heart disease. The simple reaction would be to intervene and to promote wine consumption. However, causal reasoning suggests thinking twice whether wine-reduced heart disease is a causal-effect relation or the association is confounded by other factors, say, people who drink wine are have more money and can buy better quality food, or have better quality of life in general. Counterfactual reasoning, on the other hand, often answers questions in retrospective such as "Would the outcome change if I had acted differently?".  Imagine a doctor who is treating a patient with kidney stones. The doctor is left with two choices, conventional treatment that includes open surgical procedures and a new treatment that only involves making a small puncture in the kidney. Each treatment may result in certain complications, raising the following questions:  “Would the outcome be different if the other treatment had been given to this patient.” The “if” statement is a counterfactual condition, a scenario that never happened. The doctor cannot go back in time, give a different treatment to same patient under the same exact condition. So it behooves the doctor to think about counterfactual questions in advance. Counterfactual reasoning enables us to contemplate alternative options in decision-making to possibly avoid undesired outcomes. By understating causality, we will be able to answer questions related to intervention or counterfactuals, concepts we aim to cover in the following sections.  

\subsubsection{Randomized Controlled Trials}

Due to the mentioned dilemma, a unit-level of causal effect cannot be directly observed as potential outcomes for an individual subject cannot be observed in a single universe.   Randomized controlled trials (RCT) enable us to gauge the population-level causal effect by comparing the outcomes of two groups under different treatments, while other factors are kept identical. Then, the population-level causal effect can be expressed as \textit{average causal effect}(ACE) in mathematical terms. For instance, 

\begin{equation}
    ACE = |Prob(Recovery|Treatment=Chemotherapy) - Prob(Recovery|Treatment=Placebo)|,
\end{equation}{}

 wwhere ACE is also referred as \textit{average treatment effect} (ATE).
 
In a randomized controlled trial (RCT), treatment and placebo are assigned randomly to groups that have the same characteristics (e.g., demographic factors). The mechanism  is to "disassociate variables of interest (e.g., treatment, outcome) from other factors that would otherwise affect them both"\cite{pearl2018book}.

Another factor that might greatly bias our estimation of causal effect is a century-old problem of “finding confounders” \cite{julious1994confounding,holland1983lord, greenland2001confounding}. Randomized controlled trial was firstly introduced by James Lind in 1747 to identify treatment for scurvy, and then popularized by Ronald A. Fisher in the early $20^{th}$ century. It is currently well acknowledged and considered as the golden standard to identify the true causal effect without distortions introduced by confounding.
However, randomized controlled trials are not always feasible in clinical studies due to ethical or practical concerns. For example, in a smoking-cancer medical study, researchers have to conduct randomized controlled trials to investigate if in fact smoking leads to cancer. Utilizing such trials, researchers should randomly assign participants to an experiment group where people are required to smoke and a control group where smoking is not allowed. This study design will ensure that smoking behavior is the only variable that differs between the groups, and no other variables (i.e., confounders) will bias the results. On the contrary, an observational study where we merely follow and observe the outcomes of smokers and non-smokers will be highly susceptible to confounders and can reach misleading conclusions. Therefore, the better study design would be to choose RCTs, however, it is perceived as highly unethical to ask participants to smoke in a clinical trial. 
Typically, randomized controlled trials are often designed and performed in a laboratory setting where researchers have full control over the experiment. In many real-world studies, data are collected from observations when researchers cannot intervene/randomize the independent variables. This highlights the need for a different toolkit to perform causal reasoning in such scenarios. In section 3, we will discuss how to gauge the true causal effect from observational studies that might be contaminated with confounders. In section \ref{chap:chapter_deconfound}, we will discuss how to gauge the true causal effect from observational studies that might be contaminated with confounders. 

\section{Preliminaries: structural causal models, causal graphs, and intervention with do-calculus}
\label{chap:chapter_preliminary}
In this section, we primarily introduce 3 fundamental components of causal reasoning: structural causal model (SCM), directed acyclic graphs (DAG), and intervention with do-calculus.

\subsection{Structural Causal Model}

The structural causal model (SCM), $\mathcal{M}$, is proposed by Pearl et al. \cite{pearl2009causal, hunermund2019causal} to formally describe the interactions of variables in a system. A SCM is a 4-tuple $\mathcal{M} = <U,V,F,P(u)>$ where

\begin{enumerate}
    \item $U$ is a set of background variables, exogenous, that are determined by factors outside the model.
    \item $V=\{V_1, ..., V_n\}$ is a set of endogenous variables that are determined by variables within the model.
    \item $F$ is a set of functions  $\{f_1, ..., f_n\}$ where each  $f_i$ is a mapping from $U_i \cup PA_i$ to $V_i$, where $U_i \subseteq U$ and $PA_i$ (PA is short for "parents") is a set of causes of  $V_i$. In other words, $f_i$ assigns a value to the corresponding $V_i \in V$, $v_i \la f_i(pa_i,v_i)$, for $i=1,...,n$.
    \item $P(u)$ is a probability function defined over the domain of $U$.
\end{enumerate}

Note that there are two sets of variables in a SCM, namely, exogenous variables,$U$, and endogenous variables, $V$. Exogenous variables are determined outside of the model and are not explained (or caused) by any variables inside our model. Therefore, we generally assume certain probability distributions $P(u)$ to describe the external factors. The values of endogenous variables, on the other hand, are assigned by both exogenous variables and endogenous variables. These causal mappings are captured by a set of non-parametric functions $F=\{f_1,...,f_n\}$.$f_i$ can be a linear or non-linear function to interpret all sorts of causal relations. The value assignments of endogenous variables are also referred to as data generation process (DGP) where nature assigns the values of endogenous variables.

Let us consider a toy example: in a smoking-lung cancer study, we can observe and measure the treatment variable \textit{smoking (S)} and the outcome variable \textit{lung cancer (L)}. Suppose these two factors are endogenous variables. There might be some unmeasured factors that interact with the existing endogenous variables, e.g., \textit{genotype (G)}. Then the SCM can be instantiated as,

\begin{align}
    &U = \{G\}, \quad V=\{S, L\}, \quad F=\{f_S, f_L\} \\
    &f_S: S \la f_S(G) \\
    &f_L: L \la f_L(S, G)
\end{align}

This SCM model describes that both \textit{genotype} and \textit{smoking} are direct causes of \textit{lung cancer}. Certain genotype is responsible for nicotine dependence hence explains the smoking behavior \cite{verde2011smoking,mackillop2010role}. However, no variable in this model explains variable \textit{genotype} and $G$ is an exogenous variable.

\subsection{Directed Acyclic Graph}

Every SCM is associated with a directed acyclic graph (DAG). The vertices in the graph are variables under study and causal mechanisms and processes are edges in DAG. For instance, if variable $X$ is the direct cause of variable $Y$, then there is a directed edge from $X$ to $Y$ in the graph. The previous SCM model can be visualized as follows:

\begin{figure}[ht!]
    \centering
    \includegraphics[width=70mm]{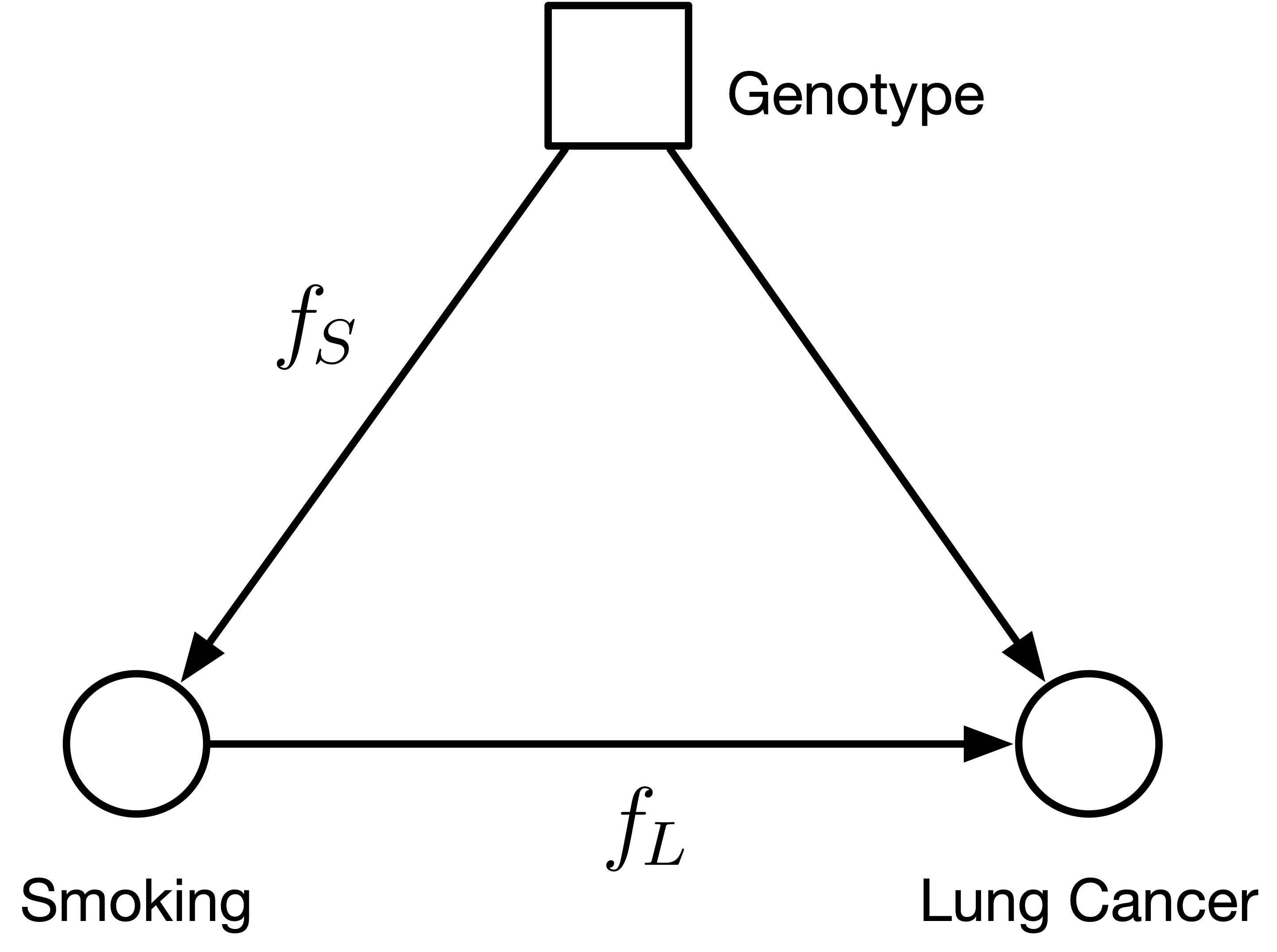}
    \caption{Graphical representation of SCM model in section 2.1. Square node denotes the exogenous variable and round nodes denote the endogenous variable. The directed edge represents the causal mechanism.}
    \label{fig:scm}
\end{figure}

Note that the graphical representation encodes the causal relations in equation 2) - 4) via a rather intuitive way. In the next section, we will show the strengths of the graphical representation when we need to study the independent relations among variables.

\subsection{Intervention with do-calculus $do(\cdot)$}

\label{sec:do_calculusx}

Do-calculus was developed by Pearl \cite{pearl2012calculus,tucci2013introduction} to gauge the effects of causal interventions \cite{pearl2012calculus,tucci2013introduction}. In the example of smoking-lung cancer, the likelihood of getting lung cancer in case of smoking, can be expressed in this conditional probability, $Prob(L=1|do(S=1))$, which describes the 

cause-effect identified in a randomized controlled trial. $L$ and $S$ are Bernoulli random variables which only take two values: 0 and 1.  $L=1$ denotes the fact of getting lung cancer, and $L=0$ represents the observation of no lung cancer. $S=1$ means that the observation of smoking whereas $S=0$ says no smoking is observed. In other words, this post-intervention distribution represents the probability of getting lung cancer $(L=1)$ when we intervene upon the data generation process by deliberately forcing participant to smoke, i.e., $do(S=1)$. Post-intervention probability distribution refers to probability terms that contain do-notation, $do(\cdotp)$. This post-intervention distribution is different from the conditional probability in an observational study: $Prob(L=1|S=1)$, which only represents the likelihood of outcome $(L=1)$ when we observe that someone smokes. This conditional probability in observational study does not entail the true causal effect as it might be biased by confounders. We will discuss the confounding issue in the next section. 

To recall, randomized controlled trials might be impractical or even unethical to conduct at times. For example, we cannot force participants to smoke in a randomized controlled experiment in order to find the cause-effect of an intervention $(do(S=1))$. Do-calculus suggests us to raise the following question instead: is it possible to estimate the post-intervention $Prob(L|do(S))$ from observational study. If we can express $Prob(L|do(S))$ in terms of the conditional probability $Prob(L|S)$ estimated in the observational study, then we can gauge the causal-effect without performing randomized controlled trials.

\subsubsection{do-calculus algebra}

Here we introduce the algebraic procedure of do calculus that allows us to bridge the gap of probability estimation between observational study and randomized controlled trials. The goal of do-calculus is to reduce the post-intervention distribution that contains the $do(\cdotp)$ operator into a set of probability distributions of $do(\cdotp)$ free. The complete mathematical proof of do-calculus can be seen in \cite{huang2012pearl,pearl2009causal}.

{\rul Prob(Y|do(X),Z,W)=Prob(Y|do(X),Z) when we observe the variable W is irrelevant to Y (possibly conditional on the other variable Z), then the probability distribution of Y will not change.}

{\rul Prob(Y|do(X),Z)=Prob(Y|X,Z) if Z is a set of variables blocking all "back-door" paths from X to Y, then $Prob(Y|do(X),Z)$ is equivalent to $Prob(Y|X,Z)$. Backdoor path will be explained shortly.  }

{\rul Prob(Y|do(X))=Prob(Y) we can remove do(X) from Prob(Y|do(X)) in any case where there are no causal paths from X to Y. If it is not feasible to express the post-intervention distribution, Prob(L|do(S)), in terms of do-notation-free conditional probabilities (e.g., Prob(L|S)) using the aforementioned rules, then randomized controlled trials are necessary to gauge the true causality.}

\subsubsection{Backdoor path and d-separation}

In rule 2, the backdoor path refers to any path between cause and effect with an arrow pointing into cause in a directed acyclic graph (or a causal graph). For example, the backdoor path between smoking and lung cancer in Figure 2 is "smoking $\la$ genotype $\ra$ lung cancer". How do we know if a backdoor path is blocked or not?

If it is not feasible to express the post-intervention distribution, $Prob(L|do(S))$, in terms of do-free conditional probability, $Prob(L|S)$, using the aforementioned rules, then randomized controlled trials are required to gauge the true causality.

 In rule 2, the backdoor path refers to the any path between cause and effect with an arrow pointing into cause in a directed acyclic graph (or a causal graph). For example, the backdoor path between smoking and lung cancer in Figure \ref{fig:scm} is "smoking $\la$ genotype $\ra$ lung cancer". How do we know if a backdoor path is blocked or not?

Judea Pearl in his book \cite{pearl2018book} introduced the concept of d-separation that tell us how to block the backdoor path. Please refer to \cite{geiger1990identifying} for the complete mathematical proof.

\begin{enumerate}[label=\alph*)]
    \item In a chain junction, A $\ra$ B $\ra$ C, conditioning on B prevents information about A from getting to C or vice versa.
    \item In a fork or confounding junction, A $\la$ B $\ra$ C, conditioning on B prevents information about A from getting to C or vice versa.
    \item In a collider, A $\ra$ B $\la$ C, exactly the opposite rules hold. The path between A and C is blocked when not conditioning on B. If we condition on B, then the path is unblocked. Bear in mind if this path is blocked A and C would be considered independent of each other.
\end{enumerate}{}

In figure 2, conditioning on genotype will block the backdoor path between smoking and lung cancer. Here, conditioning on genotype means that we only consider a specific genotype in our analysis. Blocking the backdoor between the cause and effect actually prevents the spurious correlation between them in an observational study. Please refer to the next section for more details on confounding bias.

\subsection{What is the difference between $Prob(Y=y|X=x)$ and $Prob(Y=y|do(X=x))$?}

In \cite{pearl2009causal}, Pearl et al explains the difference between the two distributions as follows, "In notation, we distinguish between cases where a variable $X$ takes a value $x$ naturally and cases where we fix $X=x$ by denoting the latter $do(X=x)$. So $Prob(Y=y|X=x)$ is the probability that $Y=y$ conditional on finding $X=x$, while $Prob(Y=y|do(X=x))$ is the probability that $Y=y$ when we intervene to make $X=x$. In the distributional terminology, $Prob(Y=y|X=x)$ reflects the population distribution of $Y$ among individuals whose $X$ value is $x$. On the other hand, $Prob(Y=y|do(X=x))$ represents the population distribution of $Y$ if everyone in the population had their $X$ value fixed at $x$".
This can be better understood with a thought experiment. Imagine that we study the association of barometer readings and weather conditions. We can express this association in terms of $Prob(Barometer|Weather)$ or $Prob(Weather|Barometer)$. Notice that correlations can be defined in both directions. However, causal relations are generally uni-directional. $Prob(Weather=rainy|Barometer=low)$ represents the probability of weather being rainy when seeing the barometer reading is low. $Prob(Weather=rainy|do(Barometer=low))$ describes the likelihood of weather being rainy after we manually set the barometer reading to low. Our common sense tells us that manually setting the barometer low would not affect the weather condition, hence, this post-intervention probability should be zero, whereas $Prob(Weather=rainy|Barometer=low)$ might not be zero.

\subsection{From Bayesian networks to Structural Causal Models.}

\begin{figure}[ht!]
    \centering
    \includegraphics[width=120mm]{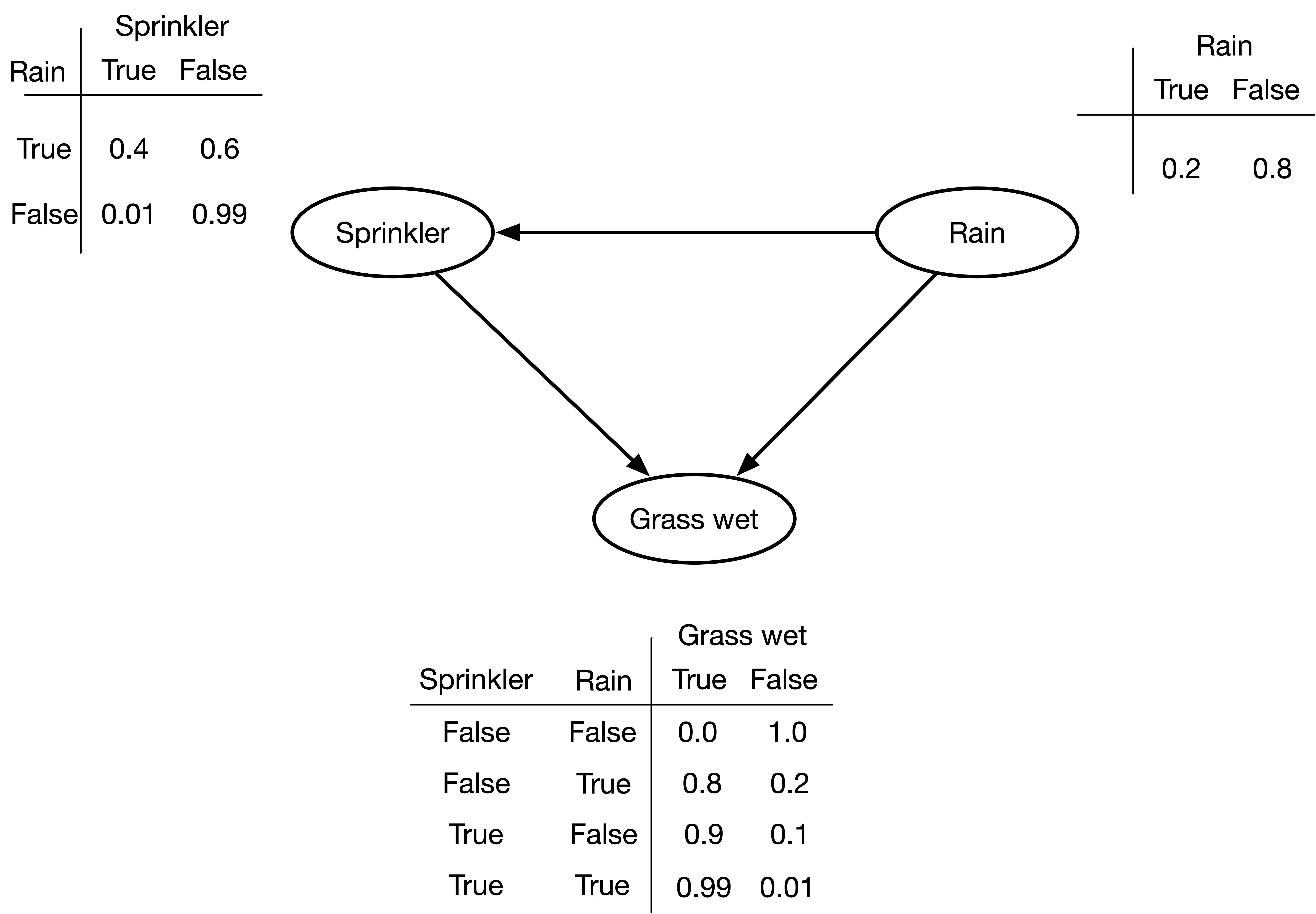}
    \caption{A simple example of Bayesian network with conditional probability tables.}
    \label{fig:bn}
\end{figure}

Some readers may raise the question: “what is the connection between structural causal models and Bayesian networks, which also aims to interpret causality from the data using DAGs?”. Firstly,  Bayesian network (also know as belief networks) was introduced by Pearl \cite{pearl1985bayesian} in 1985 as his early attempt into causal inference. A classic example  of Bayesian network is shown in Fig. Fig.\ref{fig:bn}. The nodes in Bayesian networks represent the variables of interests, and the edges between linked variables denote their dependencies, and the strength of such dependencies are quantified by conditional probabilities. The directed edges in this simple network (Fig.\ref{fig:bn}) encodes the following causal assumptions: 1) \textit{Grass wet} is true if the \textit{Sprinkler} is true or \textit{Rain} is true. 2) \textit{Rain} is the direct cause of \textit{Sprinkler} as the latter is usually off in a rainy day to preserve the water usage. We can use this probabilistic model to reason the likelihood of a cause given an effect is observed, e.g., the likelihood of a rainy day if we observe the sprinkler is on is $Prob(Rain=True |Sprinkler=True)=0.4$ as shown in the conditional probability tables in Fig.\ref{fig:bn}. 

However, “\textit{a Bayesian network is literally nothing more than a compact representation of a huge probability table. The arrows mean only that the probabilities of child nodes are related to the values of parent nodes by a certain formula (the conditional probability tables)}” \cite{pearl2018book}. On the contrary, the arrows in the structural causal models describe the underlying data generation process between linked variables (i.e., cause and effect) using a function mapping instead of conditional probability tables. If that we construct a SCM on the same example, the DAG remains unchanged, but the interpretations of the edges are different. For example, the edge of “$Rain\rightarrow Sprinkler$” indicates the function $Sprinkler \leftarrow f(Rain)$, which dictates how the effect ($Sprinkler$) would respond if we wiggle the cause ($Rain$). Note that $Sprinkler$ is the effect and $Rain$ is the cause, and the absence of the arrow “$Sprinkler \leftarrow Rain$” in the DAG says there is no such function, $Rain \leftarrow f(Sprinkler)$. Consider we would like to answer an interventional question, “What is likelihood of a rainy if we manually turn on the sprinkler, $Prob(Rain=True |do(Sprinkler=True))$?”. It is natural to choose SCMs for such questions: since we know that $Sprinkler$ is not the direct cause of $Rain$ according to the causal graph, the \textbf{rule 3} of do-calculus algebra (section 2.3.1) permits us to reduce $Prob(Rain=True|do(Sprinkler=True))$ to $Prob(Rain=True)$. That is the status of $Sprinkler$ has no impact on $Rain$. However, a Bayesian network is not equipped to answer such interventional and counterfactual questions.  The conditional probability $Prob(Sprinkler=True |Rain=True )=0.4$ only says the association between $Sprinkler$ and $Rain$ exists. Therefore, the ability to emulate interventions is one of the advantages of SCMs over Bayesian networks \cite{pearl2018book,pearl2001bayesian}.

However, Bayesian networks is an integral part of the development of causal inference framework as it is an early attempt to marry causality to graphical models.  All the probabilistic properties (e.g., local Marko property) of Bayesian network are also valid in SCMs \cite{pearl2018book,geiger1990identifying,pearl2001bayesian}.  Meanwhile, Bayesian networks also impact causal discovery research, which focuses on the identification of causal structures from data through computational algorithms \cite{heckerman1999bayesian}. 

\section{Simpson paradox and confounding variables}
\label{chap:chapter_deconfound}

\subsection*{Spurious correlations introduced by confounder}

The famous phrase "correlation does not imply causation" suggests that the observed correlation between variables A and B does not automatically entail causation between A and B. Spurious correlations between two variables may be explained by a confounder. For example, considering the following case (Fig. 4) where a spurious correlation between yellow fingernails and lung cancer is observed. One cannot simply claim that people who have yellow fingernails have higher risk of lung cancer as neither is the cause of the other. Confounding is a causal concept and cannot be expressed in terms of statistical correlation  \cite{vanderweele2013definition, greenland1999confounding}. 

\begin{figure}[ht]
    \centering
    \includegraphics[width=70mm]{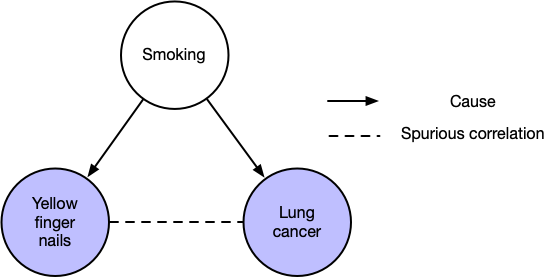}
    \caption{Smoking is a common cause and confounder for yellow finger nails and lung cancer. A spurious correlation may be observed between two groups who have yellow fingernails and lung cancer because of the third variable smoking}
    \label{fig:confounder}
\end{figure}{}

Another interesting study \cite{chocolate} reported there is a "surprisingly powerful correlation" ($rho=0.79,\ p < 0.0001$) between the chocolate consumption and the number of Nobel Laureates in a country. It is hard to believe there is any direct causal relation between these two variables in this study. This correlation might be again introduced by a confounder (e.g., advanced educational system in developed countries) that is not included in this study.

Pearl argues that \cite{pearl2009causal}: "\textit{one cannot substantiate causal claims from association alone, even at the population level. Behind every causal conclusion there must lie some causal assumptions that is not testable in an observational study}".

\subsubsection{Simpson paradox example: kidney stone}

Confounder (a causal concept) may not only introduce spurious correlations but can also generate misleading results. Table \ref{table:kidney} shows a real-life medical study \cite{charig1986comparison} that compares the effectiveness of two treatments for kidney stones. Treatment A includes all open surgical procedures while treatment B is percutaneous nephrolithotomy (which involves making only a small puncture(s) in the kidney). Both treatments are assigned to 2 groups with the same size (i.e., 350 patients). The fraction numbers indicate the number of success cases over the total size of the group.

If we consider the overall effectiveness of two treatments, treatment A (success rate$=\frac{273}{350}=0.78$) is inferior to treatment B (success rate$=\frac{289}{350}=0.83$). At this moment, we may think treatment B has higher chance of cure. However, if we compare the treatment by the size of the kidney stones, we discover that treatment A is clearly better in both groups, patients with small stones and patients with large stones. Why is the trend at the population level is reversed when we analyze treatments in sub-populations?

\begin{table}[ht]
\caption{Kidney stone treatment. The fraction numbers indicate the number of success cases over the total size of the group. Treatment B is more effective than treatment A at overall population level. But the trend is reversed in sub-populations.}

\label{table:kidney}
\centering
\begin{tabular}{c|c|c}
                                                 \hline                               & Treatment A            & Treatment B            \\ \hline
\begin{tabular}[c]{@{}c@{}}Small Stones\\ ($\frac{357}{700}=0.51$)\end{tabular} & $\frac{81}{87}=0.93$   & $\frac{234}{270}=0.87$ \\ \hline
\begin{tabular}[c]{@{}c@{}}Large Stones\\ ($\frac{343}{700}=0.49$)\end{tabular} & $\frac{192}{263}=0.73$ & $\frac{55}{80}=0.69$   \\ \hline
Overall                                                                         & $\frac{273}{350}=0.78$ & $\frac{289}{350}=0.83$\\
\hline
\end{tabular}
\end{table}

\begin{figure}[ht]
    \centering
    \includegraphics[width=50mm]{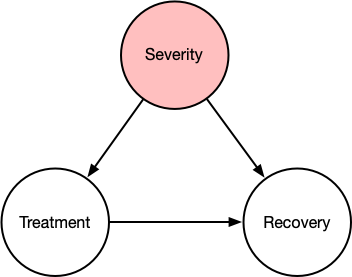}
    \caption{Observational study that has a confounder, \textit{severity}}
    \label{fig:kidney}
\end{figure}

If we inspect the table with more caution, we realize that treatments are assigned by the severity, i.e., people with large stones are more likely to be treated with method A while most of those with small stones are assigned with method B. Therefore, severity (the size of the stone) is a confounder that affects both the recovery and treatment as shown in Fig.\ref{fig:kidney}. 

{}

Ideally, we are interested in the pure causal relation of "treatment X $\ra$ recovery" without any other unwanted effects from exogenous variables (e.g., the confounder severity). We de-confound the causal relation of "treatment X $\ra$ recovery" by intervening on variable $Treatment$ and forcing its value to be either $A$ or $B$. By fixing the treatment, we can remove the effect coming from variable $Severity$ to variable $Treatment$. Note that the causal edge of "$Severity \ra Treatment$" is absent in the mutilated graphical model shown in Fig. 5. Since Severity does not affects the $Treatment$ and $Recovery$ at the same time after the intervention, it is no longer a confounder. Intuitively, we are interested in understanding if we use treatment A on all patients, what will be the recovery rate, $Prob(Recovery|do(Treatment=A))$. Similarly, what is the recovery rate, $Prob(Recovery|do(Treatment=B))$, if we use treatment B only. If the former has larger value, then treatment A is more effective; otherwise, treatment B has higher chance of cure. The notation $do(X=x)$ is a do-expression which fixes the value of $X=x$. Note that the probability $Prob(Recovery|do(Treatment))$ marginalizes away the effect of severity by $Prob(Recovery|do(Treatment))=Prob(Recovery|do(Treatment)$,$Severity=treatmentA)+Prob(Recovery|do(Treatment)$,$Severity=treatmentB)$. Essentially, we are computing the causal effects of "treatment A→ recovery" and "treatment B→ recovery":

\begin{figure}[ht]
    \centering
    \includegraphics[width=50mm]{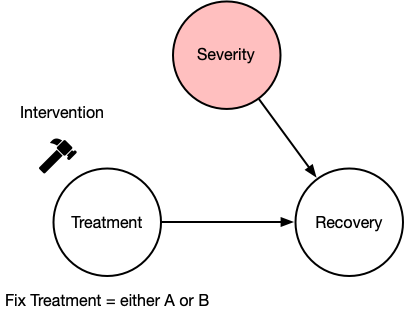}
    \caption{We simulate the intervention in the form of a mutilated graphical mode. The causal effect $Prob(Recovery|do(Treatment))$ is equal to the conditional probability $Prob(Recovery|Treatment)$ in this mutilated graphical model.}
    \label{fig:multilated}
\end{figure}{}

\begin{align}
    &Prob(R=1 | do(T=A)) \\
    &= \sum_{s \in \{small, large\}}Prob(R=1, S=s | do(T=A)) \text{\quad( \textit{law of total probability})}\\
    &= \sum_{s \in \{small, large\}}Prob(R=1 |S=s, do(T=A)) Prob(S=s|do(T=A)) \text{\quad (definition of conditional probability)}\\
    &= \sum_{s \in \{small, large\}}Prob(R=1 |S=s, do(T=A)) Prob(S=s) \text{\quad (rule \#3 in \textit{do-calculus}, see section \ref{sec:do_calculusx}})\\
    &= \sum_{s \in \{small, large\}}Prob(R=1 |S=s, T=A) Prob(S=s) \text{\quad (rule \#2 in \textit{do-calculus}, see section \ref{sec:do_calculusx})}\\
    &= Prob(R=1|S=small,T=A)Prob(S=small) + Prob(R=1|S=large,T=A)Prob(S=large)\\
    &= 0.93 \times 0.51 + 0.73 \times 0.49\\
    &= 0.832
\end{align}

Similarly, we can compute,

\begin{align}
    &Prob(R=1 | do(T=B)) \\
    &= \sum_{s \in \{small, large\}}Prob(R=1, S=s | do(T=B)) \text{\quad( \textit{law of total probability})}\\
    &= \sum_{s \in \{small, large\}}Prob(R=1 |S=s, do(T=B)) Prob(S=s|do(T=B)) \text{\quad (definition of conditional probability)}\\
    &= \sum_{s \in \{small, large\}}Prob(R=1 |S=s, do(T=B)) Prob(S=s) \text{\quad (rule \#3 in \textit{do-calculus}, see section \ref{sec:do_calculusx}})\\
    &= \sum_{s \in \{small, large\}}Prob(R=1 |S=s, T=B) Prob(S=s) \text{\quad (rule \#2 in \textit{do-calculus}, see section \ref{sec:do_calculusx})}\\
    &= Prob(R=1|S=small,T=B)Prob(S=small) + Prob(R=1|S=large,T=B)Prob(S=large)\\
    &= 0.87 \times 0.51 + 0.69 \times 0.49\\
    &= 0.782
\end{align}

Now we know the causal effects of $Prob(Recovery|do(Treatment=A))=0.832$ \\
and $Prob(Recovery|do(Treatment=B))=0.782$. Treatment A is clearly more effective than Treatment B. The results also align with our common sense that open surgery (treatment A) is expected to be more effective. A more informative interpretation of the results is that the difference of the two causal effects denotes the fraction of the population that would recover if everyone is assigned with treatment A compared to the other procedure. Recall that we have the opposite conclusion if we read the "effectiveness" at population level in table \ref{table:kidney}.

\subsection{How to estimate the causal effect using intervention?}

The "interventionist" interpretation of causal effect is often described as the magnitude by which outcome $Y$ is changed given a unit change in treatment $T$. For example, if we are interested in the effectiveness of a medication in the population, we would set up an experimental study as follows: 1) We administer the drug uniformly to the entire population, $do(T=1)$, and compare the recovery rate $Prob(Y=1|do(T=1))$ to what we obtain under the opposite context $Prob(Y=1|do(T=0))$, where we keep everyone from using the drug in a parallel universe, $do(T=0)$. Mathematically, we estimate the differenceknown as ACE (defined in section 1.3),

\begin{equation}
    ACE = Prob(Y=1|do(T=1)) - Prob(Y=1|do(T=0))
\end{equation}

 "\textit{A more informal interpretation of ACE here is that it is simply the difference in the fraction of the population that would recover if everyone took the drug compared to when no one takes the drug}" \cite{pearl2016causal}. The question is how to estimate the intervention distribution with the do operator, $Prob(Y|do(T))$. We can utilize the following theorem to do so,

{\theorem \textbf{The causal effect rule }Given a graph G in which a set of variables PA are designated as the parents of X, the causal effect of X on Y is given by}

\begin{equation}
Prob(Y=y|do(X=x)) = \sum_{z}Prob(Y=y|X=x, PA=z)Prob(PA=z)
\end{equation}

If we multiply and divide the right hand side by the probability $Prob(X=x|PA=z)$, we get a more convenient form:

\begin{equation}
Prob(y|do(x)) = \sum_{z}\frac{Prob(X=x, Y=y, PA=z)}{Prob(X=x, PA=z)}
\end{equation}

Now the computation of $Prob(Y|do(T))$ is reduced to the estimation of joint probability distributions $Prob(X, Y, PA)$ and $Prob(X, PA)$, which can be directly computed from the corresponding observational dataset. Please refer to \cite{murphy2012machine} for details on probability distribution estimation.

\section{External validity and transportability of machine learning models.}
\label{chap:chapter_transportability}

In the era of big data, we diligently and consistently collect heterogeneous data from various studies. For example, data collected from different experimentational conditions, underlying population, locations, or even different sampling procedures. In short, the collected data are messy, and rarely serves our inferential goal. Our data analysis should account for these factors. "\textit{The process of translating the results of a study from one setting to another is fundamental to science. In fact, scientific progress would grind to a halt were it not for the ability to generalize results from laboratory experiments to the real world}" \cite{pearl2018book}. We initially need to take a better look at heterogenous datasets. 

\subsection{How to describe the characteristics of heterogeneous datasets?}

\begin{figure}[ht]
    \centering
    \includegraphics[width=90mm]{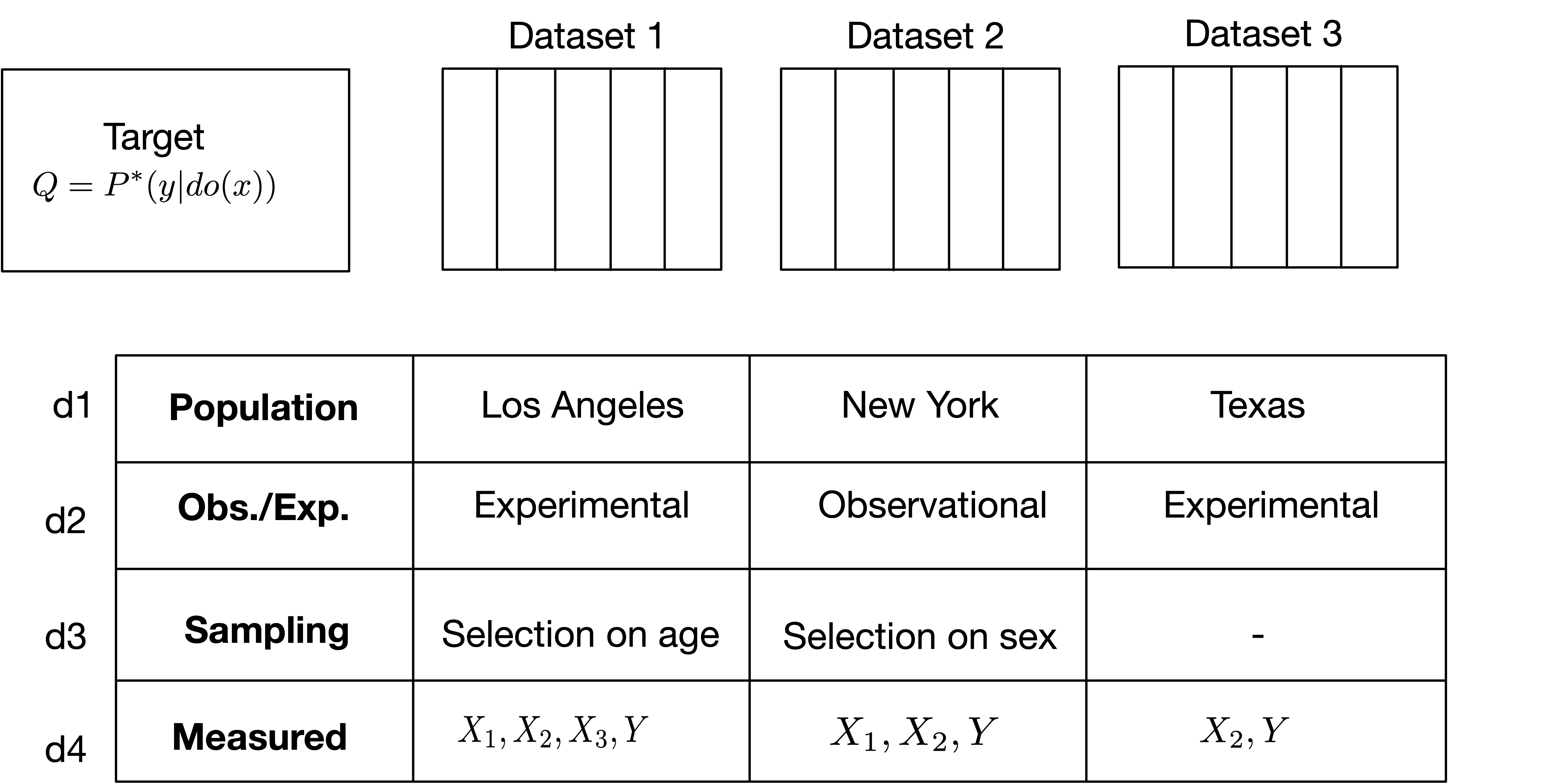}
    \caption{Heterogeneous datasets can vary on the dimensions ($d1,d2,d3,d4$) shown above. Suppose we are interested in the causal effect $X \ra Y$ in a study carried out in Texas, and we have the same causal effect studied in Los Angeles and New York. This table exemplifies the potential differences between the datasets \cite{EBtalkColumbia}}.
    \label{fig:heterogenous_dataset}
\end{figure}{}

Big data empowers us to conduct a wide spectrum of studies and to investigate the analytical results. We normally incline to incorporate or transfer such results to a new study. It naturally raises the question: under what circumstances can we transfer the existing knowledge to new studies that are under different conditions. Before we come up with “licenses” of algorithms that permits such transfer, it is crucial to understand how the new dataset in the target study differs from the ones in the existing studies.

Bareinboim \cite{bareinboim2016causal} summarizes the differences of heterogeneous datasets over the four dimensions shown in Fig. \ref{fig:heterogenous_dataset} \cite{bareinboim2016causal} that are certainly not enough to enumerate all possibilities in real practice, but more dimensions can be added in future research. In Fig. \ref{fig:heterogenous_dataset}, 
\begin{itemize}
    \item d1) datasets may vary on the study population;
    \item d2) datasets may vary in study design. For instance, the study in Los Angeles is an experimental study under laboratory setting, while the study in New York is an observational study in real world;
    \item d3) datasets may vary in collection process. For instance, dataset 1 may suffer from selection bias on variable age; for example, if the subjects recruited in study 1 are contacted only using landlines, the millennials probably are excluded in the study as they prefer mobile phones;
    \item d4) Studies might also take measurements on different sets of variables.
\end{itemize}{}

\subsection{Selection bias}
 
 Selection bias is caused by preferential exclusion of data samples \cite{bareinboim2012controlling}. It is a major obstacle in validating statistical results, and it can hardly be detected in either experimental or observational studies.

\subsubsection{COVID-19 example}

During the COVID-19 pandemic crisis, a statewide study reported that 21.2\% of New York City residents have been infected with COVID-19 \cite{NYC}. The study tested 3,000 New York residents statewide at grocery and big-box stores for antibodies that are to indicate whether someone has had the virus. Cassie Kozyrkov \cite{NYC_selection_bias} argues the study might be contaminated with selection bias. The hypothesis notes that the cohort in the study is largely skewed towards the group of people who are high risk-takers/less cautious and have had the virus. The large portion of the overall population may include people who are risk-averse and cautiously stay home; these people were excluded from the research samples. Therefore, the reported toll (i.e., 21.2\%) is likely to be inflated. Here, we try to investigate a more generic approach to spot-on selection bias issues.

Causal inference requires us to make certain plausible assumptions when we analyze data. Data sets are not always complete, that is, it does not always tell the whole story. The aresults of the analyses (e.g., spurious correlation) from data alone can be often very misleading. You may recall the example of smokers who may develop lung cancer and have yellow fingernails. If we find this association (lung cancer and yellow fingernails) to be strong, we may come to the false conclusion that one causes the other.

Back to our COVID-19 story, what causal assumptions can we make in the antibody testing study? Consider Fig. 8 in which each of the following assumptions represent an edge.

\begin{enumerate}[label=\roman*)]
    \item We know that the antibody will be discovered if we do the related test.
    \item People who have had COVID-19 and survived would generate an antibody for that virus.
    \item Risk taking people are more likely to go outdoor and participate in the testing study.
    \item In order to highlight the difference between the sample cohort and the overall population in the graph, Bareinboim \cite{bareinboim2016causal,bareinboim2012controlling} proposed a hypothetical variable S (standing for “selection”). The variable bounded in the square in Fig. 8 stands for the characteristics by which the two populations differ. You can also think of S as the inclusion/exclusion criteria. Variable S have incoming edges from variables “risk taking” and “carried virus”. This means that the sample cohort and overall population differ in these two aspects.
\end{enumerate}{}

\begin{figure}[ht]
    \centering
    \includegraphics[width=70mm]{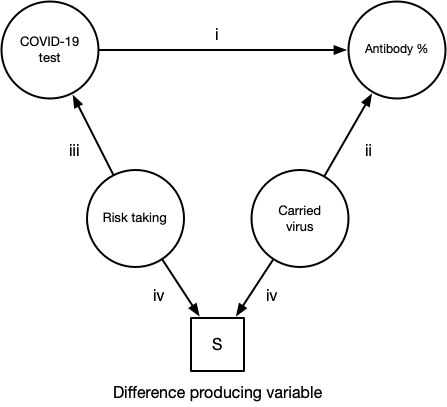}
    \caption{Graphical model that illustrates the selection bias scenario. Variable S (squared shape) is a difference producing variable, which is a hypothetical variable that points to the characteristic by which the two populations differ.}
    \label{fig:covid-19-sb}
\end{figure}{}
   
 Now we have encoded our assumptions into a transparent and straightforward diagram. You may wonder why we go through all the hassle to make this graphical diagram? What value does it add to our identification of selection bias or even debiasing procedure?
 
Let us begin with the question of how it helps us to find the common pattern/principle of identifying selection bias?

\subsubsection{Identify selection bias with causal graph}

First, a couple of quick tips in identifying selection bias: 1) find any collider variable on the backdoor path between the cause variable and the effect variable, 2) selection bias occurs when your data collection process is conditioning on these collider variables \cite{bareinboim2012controlling}. Note that these tips are only sufficient but not necessary conditions in finding selection bias. By conditioning we mean considering only some of the possibilities a variable can take and not all.

The backdoor path in tip 1) refers to any path between cause variable (COVID testing) and effect variable (antibody \%) with an arrow pointing into cause variable (\cite{pearl2018book} , page 158). In our example, the only backdoor path is “COVID-19 test $\la$ risk taking $\ra$ S $\la$ carried virus $\ra$ antibody \%”. Spurious correlation will be removed if we block every backdoor path.

We observe that there are 3 basic components on the backdoor path: 1) a fork “COVID-19 test $\la$ high risk $\ra$ S”; 2) a collider “high risk $\ra$ S $\la$ carried”; 3) another fork “S $\la$ carried virus $\ra$ antibody \%”. Now we notice that this backdoor path is blocked if we do not condition on variable S in the collider (rule c). If we condition on variable S (e.g., set S=\{high risk, have had virus\}), the backdoor path will be opened up and spurious correlation is introduced in your antibody testing results.

Now we come to a realization that if we condition on collider on the backdoor path between cause $\ra$ effect and open up the backdoor path, we will encounter the selection bias issue. With this conclusion, we can quickly identify if our study has a selection bias issue given any causal graph. The procedures of this identification can also be automated when graph is complex. So, we offload this judgement to algorithms and computers. Hopefully, you are convinced at this point that using graphical model is a more generic and automated way of identifying selection bias.

\subsubsection{Unbiased estimation with do-calculus}

Now we are interested in the estimation of cause effect between “COVID-19 test $\ra$ Antibody \%”, $P(Antibody|do(test))$. In other words, the causal effect represents the likelihood of antibody discovery if we test everyone in the population. Our readers may wonder at this point the do-calculus makes sense, but how would we compute and remove the do-operator? Recall the algebric of do-calculus introduced in section \ref{sec:do_calculusx}. Let us compute $P(Antibody|do(test))$ as follows,

\begin{align}
    &P(Antibody|do(test))\\
    &=P(Antibody|do(test),\{\}) \quad \text{\# Condition on nothing, an empty set.}\\
    &=P(Antibody|test,\{\}) \quad \text{\parbox{12cm}{\# The backdoor path in Fig.\ref{fig:covid-19-sb} is blocked naturally if we condition on nothing, \{\}. According to rule b) of do-alculus, we can safely remove the \textit{do} notation.}}\\
    &=P(Antibody|test)\\
    &=\sum_{\substack{i \in \{high, low\} \\ j \in \{true, false\}}}P(Antibody,risk=i,virus=j|test) \quad \text{\# Law of total probability.}\\
    &=\sum_{\substack{i \in \{high, low\} \\ j \in \{true, false\}}}P(Antibody|test,risk=i,virus=j)P(risk=i,virus=j|test) \quad \text{\parbox{8cm}{\# Definition of conditional probability.}}\\
\end{align}{}

The last step of the equation show four probability terms measured in the study required to have an unbiased estimation. If we assume close-world (risk=\{high, low\}, virus=\{true, false\}), it means we need to measure every stratified group. Now we have seen that do-calculus can help us identify what pieces we need in order to recover the unbiased estimation.

\subsection*{Model transportability with data fusion.}

Transferring the learned knowledge across different studies is crucial to scientific endeavors. Consider a new treatment that has shown effectiveness for a disease in an experimental/laboratory setting. We are interested in extrapolating the effectiveness of this treatment in a real-world setting. Assume that the characteristics of the cohort in the lab setting is different from the overall population in real-world, e.g., age, income, etc. Direct transfer of existing findings into new setting will result in biased inference/estimation of the effectiveness of the drug. Certainly, we can recruit a new cohort that is representative of the overall population and study whether the effectiveness of this treatment is consistent. However, if we could use the laboratory findings to infer our estimation goal in the real-world, this would reduce cost of repetitive data collection and model development. Let us consider a toy example of how causal reasoning could help with data fusion.

\subsubsection{A toy example of data fusion with causal reasoning}

Imagine we developed a new treatment for a disease, and we estimated the effectiveness of the new drug for each age group in a randomized controlled experimental setting. Let $P(Recovery | do(Treatment), Age)$ denote the drug effect at each age group. We wish to generalize the lab results to a different population. Assume that the study cohort in lab setting and the target population are different and the differences are explained by a variable $S$ as shown in Fig. 8. Meanwhile, we assume these causal effects of each specific age group are invariant across populations. We are interested in gauging the drug effect in the target population ($S=s^*$), and the query can be expressed as $P(Recovery=True | do(Treatment=True), S=s^*)$. Then the query can be solved as follows:

\begin{figure}[ht]
    \centering
    \includegraphics[width=75mm]{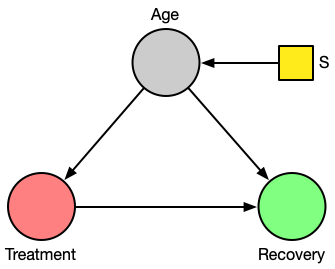}
    \caption{An toy example shows how to transfer existing inference in a lab setting to another population where the difference is the age, denoted by a hypothetical difference variable S (in yellow)}
    \label{fig:data_fusion}
\end{figure}

\begin{align}
    Query &= P(Recovery|do(Treatment), S=s^*)\\
    &= \sum_{age}P(Recovery|do(Treatment), S=s^*, Age=age)P(Age=age|do(Treatment), S=s^*)\\
    &= \sum_{age}P^*(Recovery|do(Treatment), Age=age)P^*(Age=age)
\end{align}

In the last step of the equation, $P^*(Recovery|do(Treatment), Age=age)$ is the effect we discovered through various experimental studies, and it is invariant across study cohorts. Hence, $P^*(Recovery|do(Treatment), Age=age) = P(Recovery|do(Treatment), Age=age)$. $P^*(Age=age)$ is the age distribution in the new population, and is a shorthanded for $P(Age=age|S=s^*)$. We realize that to answer the query, we just need to compute the summation of the experimental findings weighted by the age distribution in the target population.

For example, we assume that we have discovered the effectiveness of the new treatment on different age groups through some experimental studies. The effectiveness is expressed as follows,

\begin{align}
    &P(Recovery|do(Treatment), Age<10) = 0.1 \\
&P(Recovery|do(Treatment), Age=10 \sim 20) = 0.2 \\
&P(Recovery|do(Treatment), Age=20 \sim 30) = 0.3 \\
&P(Recovery|do(Treatment), Age=30 \sim 40) = 0.4 \\
&P(Recovery|do(Treatment), Age=40 \sim 50) = 0.5 \\
\end{align}

The age distribution in our target population is as follows,

\begin{itemize}
\item group1 : Age < 10 $Pb^*(Age<10) = 1/10$
\item  group2 : Age = 10 $\sim$ 20 $P^*(10 \leq Age < 20) = 2/10$
\item  group3 : Age = 20 $\sim$ 30 $P^*(20 \leq Age < 30) = 4/10$
\item  group4 : Age = 30 $\sim$ 40 $P^*(30 \leq Age < 40) = 2/10$
\item  group5 : Age = 40 $\sim$ 50 $P^*(40 \leq Age < 50) = 1/10$
\end{itemize}

According to equation 5)-7), the effectiveness in the target population should be computed as,

\begin{align}
        P(recovery|do(treatment), S=s^*) &= \sum_{age}P(recovery|do(treatment),age)P^*(age)\\
        &= 0.1 * \frac{1}{10} + 0.2 * \frac{2}{10} + 0.3 * \frac{4}{10} + 0.4 * \frac{2}{10} + 0.5 * \frac{1}{10} = 0.03
\end{align}

\section{Learn from missing data using causal inference.}
\label{chap:chapter_missingData}

\subsection*{Introduction}

Missing data occurs when the collected values are incomplete for certain observed variables. Missingness might be introduced in a study for various reasons: for examples, due to sensors that stop working because the run out of battery; Data collection is done improperly by researchers; Respondents refuse to answer some survey questions that may reveal their private information (e.g., income, disability). Missing data issue is inevitable in some scenarios. In a smoking-cancer medical study, it is perceived highly unethical to ask participants to smoke in order to test the hypothesis of smoking leading to lung cancer. 

Typically, building machine learning predictors or statistical models with missing data may expose ourselves to the following risks: a) the partially observed data may bias our inference models, and the study outcomes may largely deviate from the true value \cite{zhang2019causal}, b) the reduced sample size may lose the statistical power to provide any informative insights \cite{ryder2011advantage}, c) this technical impediment might also cause severe predictive performance degradation as most of the machine learning models assume datasets are complete when making inferences. 

\begin{figure}
    \centering
    \includegraphics[width=90mm]{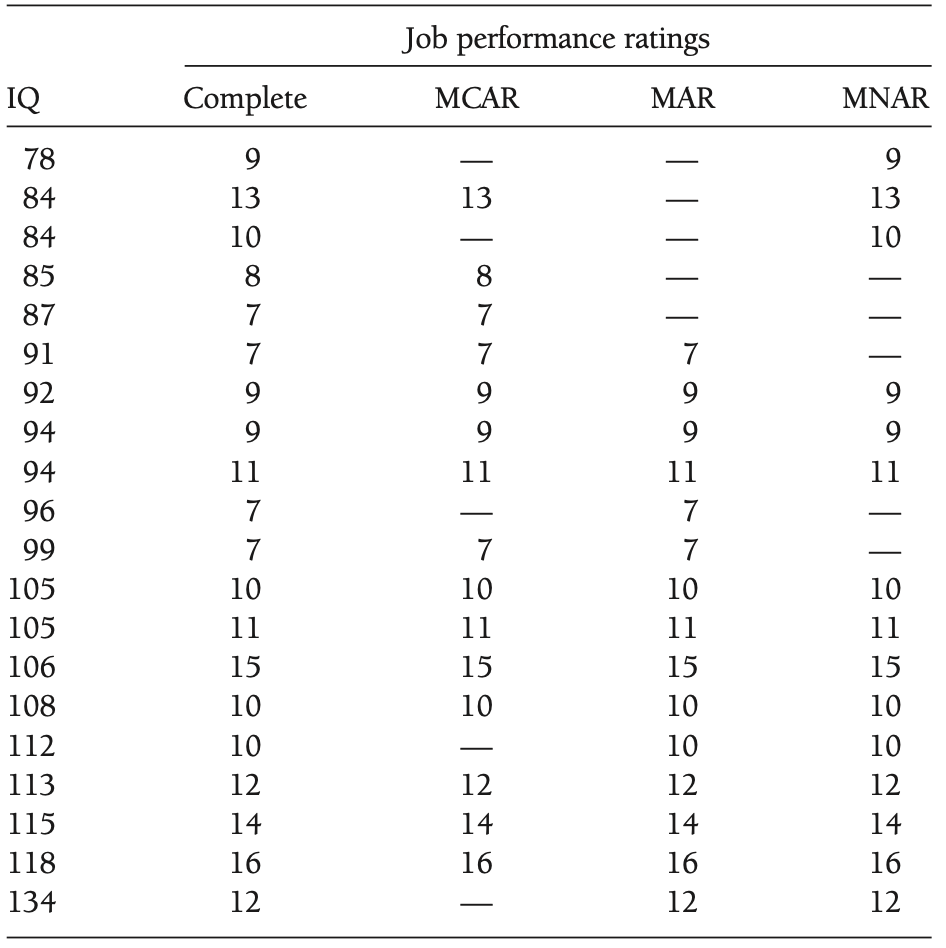}
    \caption{Example source \cite{enders2010applied}. Job performance ratings is a partially observed variable, and variable IQ is a completely observed variable without any missingness. The second column shows the complete ratings. The 3rd/4th/5th columns show the observed ratings under MCAR/MAR/MNAR conditions, respectively.}
    \label{fig:missingDataExample}
\end{figure}

Extensive research endeavors have been dedicated to this issue. List-wise deletion or mean value substitutions are commonly used in dealing with missing data because of their simplicity. However, these naive methods fail to account for the relationships between the missing data and the observed data. Thus, the interpolations usually deviate from the real values by large. Rubin et al. introduce the concept of missing data mechanism which is widely adopted in the literature \cite{rubin1976inference}. This mechanism classifies missing data into tree categories: 
\begin{itemize}
    \item Missing completely at random (\textbf{MCAR}): the observed data are randomly drawn from the complete data. In other words, missingness is unrelated to other variables or itself. For example, in Fig. \ref{fig:missingDataExample}, job performance ratings is a partially observed variable, and variable IQ is complete without any missingness. The MCAR column shows that the missing ratings are independent of IQ values and itself. 
    \item Missing at random (\textbf{MAR}): the missing values of the partially observed variable dependents on other measured variables. For example, in Fig. \ref{fig:missingDataExample}, the MAR column shows that the missing ratings are associated with low IQs. 
    \item Missing not at random (\textbf{MNAR}): MNAR includes scenarios when data are neither MCAR nor MNAR. For example, in Fig. \ref{fig:missingDataExample}, the MNAR column shows that the missing ratings are associated with itself. i.e., low job performance ratings ($ratings < 9$) are missing. 
\end{itemize}

Pearl et al. demonstrate that theoretical performance guarantee (e.g. convergence and unbiasedness) exists for inference with data that are MCAR and MAR  \cite{mohan2013graphical,pearl2013recoverability}.  In other words, we can still have bias-free estimation with even missing data. For example, in Fig. \ref{fig:missingDataExample}, assume $Y$ is the random variable that represents the job performance ratings. The expectation of job performance ratings under complete, MCAR, MNAR columns  are $\mathbb{E}^{Complete}[Y]=10.35$, $\mathbb{E}^{MCAR}[Y]=10.60$, $\mathbb{E}^{MNAR}[Y]=11.40$, respectively. It can be easily verified that bias of $Bias_{MCAR}=|\mathbb{E}^{Complete}[Y]-\mathbb{E}^{MCAR}[Y]|=0.25$  is less than $Bias_{MNAR}=|\mathbb{E}^{Complete}[Y]-\mathbb{E}^{MNAR}[Y]|=1.05$. As the size of the dataset grows, $\mathbb{E}^{MCAR}[Y]$ will converge to the real expectation value $\mathbb{E}^{Complete}[Y]$, i.e., $Bias_{MCAR}=0$. However, since the MNAR mechanism dictates that low ratings ($Y < 9$) are inherently missing from our observations, we cannot have an bias-free estimation, regardless of the sample size, if we make no assumptions of the missing mechanism. We can notice that the observed data are governed by the missing mechanism (or data generation process). Therefore, missing data issue is inherently a causal inference problem \cite{pearl2013recoverability,mohan2015missing}. Details of the causal perspective to missing data can be referred to section \ref{section:causalPerspective}.

Most statistical techniques proposed in the literature for handling missing data assume that data are MCAR or MAR \cite{yoon2018gain, buuren2010mice, deng2016multiple, schafer2002missing}. For example, expectation maximum likelihood algorithm is generally considered superior to other conventional methods (e.g. list-wise or pair wise deletion) when the data are MCAR or MAR \cite{enders2010applied}. Moreover, it provides theoretical guarantee (e.g. unbiasedness or convergence) \cite{little2019statistical} under MCAR or MAR assumptions. However, when it comes to MNAR, estimations with the conventional statistical techniques will mostly be biased. Mohan et al. report that we can achieve unbiased estimation in MNAR scenario under certain constraints using causal methods\cite{mohan2013graphical, pearl2013recoverability, mohan2015missing}.

\subsection*{Causal Perspective}
\label{section:causalPerspective}
In this section, we briefly explore and discuss the causal approaches proposed by Karthika Mohan and Judea Pearl in dealing with missing data\cite{mohan2013graphical, pearl2013recoverability, mohan2015missing}. Firstly, we introduce the concept of causal graph representation for missing data -- missing graph(s) (m-graph(s) for short). Then we introduce the definition of recoverability, a formal  definition of unbiased estimation with missing data. Next we discuss under what conditions we can achieve recoverability. At last, we identify the unsolved problems with data that are MNAR. 

\subsubsection*{Preliminary on missing graphs}

\begin{figure}
    \centering
    \includegraphics[width=160mm]{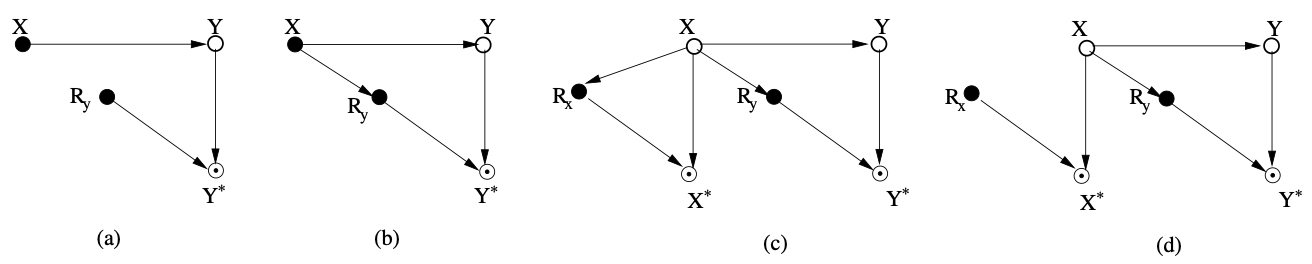}
    \caption{m-graphs for data that are: (a) MCAR, (b) MAR, (c) \& (d) MNAR; Hollow and solid circles denote partially and fully observed variables respectively \cite{mohan2013graphical}}
    \label{fig:missing_graph}
\end{figure}{}

Let $G(\mathbb{V},E)$ be the causal graph (a DAG) where $\mathbb{V}=V \cup U \cup V^{*} \cup \mathbb{R}$. $V$ denotes the set of observable nodes, which represents observed variables in our data. These observable nodes can be further grouped into fully observable nodes, $V^{obs}$, and partially observable nodes, $V^{mis}$. Hence, $V = V^{obs} \cup V^{mis}$. $V^{obs}$ denotes the set of variables that have complete values whereas $V^{mis}$ denotes the set of variables that are missing at least one data record. Each partially observed variable $v_i \in V^{mis}$ has two auxiliary variables $R_{v_i}$ and $V^{*}_i$, where $V^{*}_i$ is a proxy variable that is actually observed, and $R_{v_i}$ denotes the causal mechanism responsible for missingness of $V^{*}_i$,

\begin{equation}
    v^*_i=f(r_{v_i}, v_i) =\begin{cases}
                                v_i,  & \text{if } r_{v_i}=0\\
                                missing,  & \text{if } r_{v_i}=1
                            \end{cases}
\end{equation}

In missing graphs, $R_{v_i}$ can be deemed as a switch that dictates the missingness of its proxy variable, $V^*_i$. For example, in Figure \ref{fig:missing_graph} (a), $X$ is a fully observable node which has no auxiliary variables. Partially observable node $Y$ is associated with $Y^*$ and $R_{Y_i}$. $Y^*$ is the proxy variable that we actually observe on $Y$,  and $R_y$ masks the values of $Y$ by its underlying missingness mechanism (e.g., MCAR, MAR, MNAR). $E$ is the set of edges in m-graphs, and  $U$  is the set  of unobserved variables (latent variables). For  example, in a recommender system,  the purchase interest of a user can not be measured nor observed. Hence it is an unobserved variable. The toy example in figure \ref{fig:missing_graph} is not involved with any latent variable. 

We can cast the classification of missingness  mechanisms (e.g., MCAR, MAR, MNAR) onto m-graphs as depicted in Fig. \ref{fig:missing_graph}. 

\begin{itemize}
    \item Fig. \ref{fig:missing_graph} a) shows the MCAR case where $R_y \indep X$ \footnote[1]{This (conditional) independence is directly read off from causal graphs using \textit{d-separation} technique}. The on-off status of $R_y$ is solely determined by coin-toss. Bear in mind that the absence of an edge between two vertices in causal graph is a strong constraint which represents there is no relation between them. The criterion of deciding if a m-graph represents MCAR is $\mathbb{R}  \indep (V^{obs} \cup V^{mis} \cup U)$.
    \item Fig. \ref{fig:missing_graph} b) shows the MAR case where $R_y \indep Y | X$ \footnotemark[1]. $R_y$ depends on the fully observed variable $X$. The criterion of deciding if a m-graph represents MAR is $\mathbb{R}  \indep (V^{mis} \cup U) | V^{obs}$.
    \item Fig. \ref{fig:missing_graph} c)\&d) show the MNAR cases where neither of the aforemetioned criterions holds.
\end{itemize}

It is a clear advantage that we can directly read the missingness mechanism from the m-graphs using \textit{d-seperation} \footnote[2]{Watch this video on d-separation: \url{https://www.youtube.com/watch?v=yDs_q6jKHb0}} without conducting any statistical test.

\subsection*{Recoverability}

Before we can discuss under what conditions we can achieve bias-free estimation, we shall firstly introduce the definition of recoverability.

\begin{defi}
\textbf{Recoverability} \cite{mohan2013graphical}. Given a m-graph G, and a target query relation Q defined on the variables in V, Q is said to be recoverable in G if there exists an algorithm that produces a consistent estimate of Q for every dataset D such that P(D) is 1)  compatible with G and 2) strictly positive over complete cases, i.e., $P(V^{obs}, V^{mis}, \mathbb{R}=0) >0$.
\end{defi}

The definition in the original paper \cite{mohan2013graphical} may be a bit obscure at first. To my understanding, a query (e.g. what is the value of joint probability, $Prob(X,Y)$ in Fig. \ref{fig:missing_graph}) is recoverable if there exists an algorithm that computes $Prob(X,Y)$ with the observed values of $X, Y^*$. Then this algorithm is referred as a "consistent estimator" which gives unbiased inference. We will see some examples in this section later.

\begin{coro}\cite{mohan2013graphical}.
A query relation Q is recoverable in G if and only if Q can be expressed in terms of the probability P(O) where $O = {R,V^*,V^{obs}}$ is the set of observable variables in G. In other words, for any two models $M_1$ and $M_2$ inducing distribution $P^{M_1}$ and $P^{M_2}$ respectively, if $P^{M_1}(O)=P^{M_2}(O)>0$ then $Q^{M_1}=Q^{M_2}$.
\end{coro}

\subsubsection*{Recoverability when data are MCAR}

\begin{eg}\cite{mohan2013graphical}.
Let X be the treatment and Y be the outcome as depicted in the m-graph in Fig. \ref{fig:missing_graph} a). Let it be the case that we accidentally delete the values of Y for a handful of samples, hence $Y \in V_m$. Can we recover P(X,Y)?\\

Yes, P(X,Y) under MCAR is recoverable. We know that $R_y \indep (X, Y)$ \footnotemark[1] holds in Fig. \ref{fig:missing_graph} a). Thus, $P(X,Y) = P(X,Y|R_y) = P(X,Y|R_y=0)$. When $R_y=0$, we can safely replace $Y$ with $Y^*$ as $P(X,Y) = P(X,Y^*|R_y=0)$. Note that $P(X,Y)$ has been expressed in terms of the probability ($P(X,Y^*|R_y=0)$) we can compute using observational data. Hence, we can recover P(X,Y) with no bias. 
\end{eg}

\subsubsection*{Recoverability when data are MAR}

\begin{eg}\cite{mohan2013graphical}.
Let X be the treatment and Y be the outcome as depicted in the m-graph in Fig. \ref{fig:missing_graph} b). Let it be the case that some patients who underwent treatment are not likely to report the outcome, hence $X \in R_y$. Can we recover P(X,Y)?\\

Yes, P(X,Y) under MAR is recoverable. We know that $R_y \indep Y | X$ \footnotemark[1] holds in Fig. \ref{fig:missing_graph} b). Thus, $P(X,Y) = P(Y|X)P(X) = P(Y|X, R_y)P(X) = P(Y|X, R_y=0)P(X)$. When $R_y=0$, we can safely replace $Y$ with $Y^*$ as $P(X,Y) = P(Y*|X, R_y=0)P(X)$. Note that $P(X,Y)$ has been expressed in terms of the probability ($P(Y*|X, R_y=0)P(X)$) we can compute using observational data. Hence, we can recover P(X,Y) with no bias. 
\end{eg}

\subsubsection*{Recoverability when data are MNAR}

\begin{eg}\cite{mohan2013graphical}.
 Fig. \ref{fig:missing_graph} d). depicts a study where (i) some units who underwent treatment (X=1) did not report the outcome Y, and (ii) we accidentally deleted the values of treatment for a handful of cases. Thus we have missing values for both X and Y which renders the dataset MNAR. Can we recover P(X,Y)?\\

Yes, P(X,Y) in d) is recoverable. We know that $X \indep R_x$ and $(R_y \cup R_y) \indep Y | X$ \footnotemark[1] holds in Fig. \ref{fig:missing_graph} d). Thus, $P(X,Y) = P(Y|X)P(X) = P(Y|X, R_y)P(X) = P(Y^*|X^*, R_y=0, R_x=0)P(X^*|R_x=0)$. Note that $P(X,Y)$ has been expressed in terms of the probability ($P(Y^*|X^*, R_y=0, R_x=0)P(X^*|R_x=0)$) we can compute using observational data. Hence, we can recover P(X,Y) with no bias. 
\end{eg}

In the original paper \cite{mohan2013graphical}, $P(X,Y)$ is not recoverable in Fig 2. c). Mohan et al. provide a theorem (see Theorem 1 in \cite{mohan2013graphical}) which states the sufficient condition for recoverability.

\subsection*{Testability}
In Figure \ref{fig:missing_graph} b), we assume that the missing mechanism $R_y$ is the causal effect of $X$, hence the arrow pointing from $X$ to $R_y$. The question naturally arises: "is our assumption/model compatible with our data?" Mohan et al. propose an approach to testify the plausibility of missing graphs from the observed dataset \cite{pearl1995testability}. 

\subsubsection*{Testability of Conditional Independence (CI) in m-graphs}

\begin{defi} \cite{pearl1995testability}
Let $X \cup Y \cup Z \subseteq V_o \cup V_m \cup R$ and $X \cap Y \cap Z \neq \emptyset$. $X \indep Y | Z$ is testable if there exists a dataset $D$ governed by a distribution $P(V_o, V^*,R)$ such that $X \indep Y|Z$ is refuted in all underlying distributions $P(V_o, V_m, R)$ compatible with the distribution $P(V_o, V^*, R)$.
\end{defi}{}

In other words, if the CIs can be expressed in terms of observable variables exclusively, then these CIs are deemed testable.

\begin{theorem}
Let $X, Y, Z \subset V_o \cup V_m \cup \mathbb{R}$ and $X \cap Y \cap Z = \emptyset.$ The conditional independence statement S: $X \indep Y | Z$ is directly testable if all the following conditions hold:

\begin{enumerate}
    \item $Y \nsubseteq (R_{X_m} \cup R_{Z_m})$. In words, Y should contain at least one element that is not in $R_{X_m} \cup R_{Z_m}$
    \item $R_{X_m} \subseteq X \cup Y \cup Z$.
    In words, the missingness mechanisms of all partially observed variables in X are contained in $X \cup Y \cup Z$
    \item $R_{Z_m} \cup R_{Y_m} \subseteq Z \cup Y$. In words, the missingness mechanisms of all partially observed variables in Y and Z are contained in $Y \cup Z$
\end{enumerate} 
\end{theorem}

\subsubsection*{Testability of CIs comprising of only substantive variables}

As for the CIs that only includes substantive variables (e.g., Fig. \ref{fig:missing_graph} (b)), it is fairly easy to see $X \indep Y$ is testable when $X,Y \in V_o$.

\subsection*{Missing data from causal perspective}

Given an incomplete dataset, our first step is to postulate a model based on causal assumptions of the underlying data generation process. Secondly, we need to determine whether the data rejects the postulated model by identifiable testable implications of that model. Last step is to determine from the postulated model if any method exists that produces consistent estimates of the queries of interests.

\section{Augmented machine learning with causal inference.}
\label{chap:chapter_causal_ml}

Despite the rising popularity of causal inference research, the route from machine learning to artificial general intelligence still remains uncharted. Strong artificial intelligence aims to generate artificial agents with the same level of intelligence as human beings. The capability of thinking causally is integral for achieving this goal \cite{ford2018architects}. In this section we try to describe how to augment machine learning models with causal inference.

\subsection{Reinforcement learning with causality}

Smart agents not only passively observe the world but are also expected to actively interact with their surroundings and to shape the world. Let us consider case study in recommender systems where we augment a reinforcement learning (RL) agent with causality.

\begin{figure}[ht]
    \centering
    \includegraphics[width=70mm]{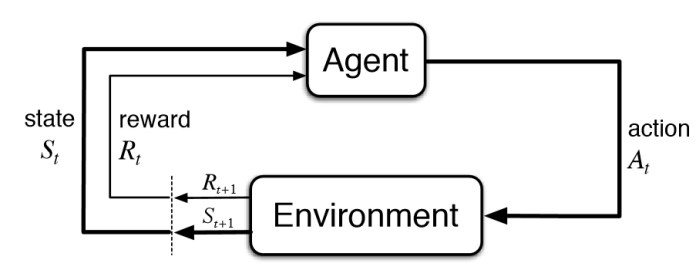}
    \caption{A graphical illustration of the interaction between a reinforcement learning agent and its surrounding environment.}
    \label{fig:confounder}
\end{figure}{}

A reinforcement learning agent typically interacts with environment as follows: choose the available \textbf{actions} from the observed \textbf{state}, and collect the \textbf{reward} as the results of those actions. The goal of a reinforcement learning agent is to maximize the cumulative reward by choosing the optimal action in every interaction with the environment. Reinforcement learning (RL) has been widely adopted in recommender systems such as Amazon or eBay platforms in which item are being suggested to users based on learning their past purchases. Let us assume that these e-commerce recommender systems are implemented as reinforcement learning agents. Given a collection of recommended items, a reward is recorded when a purchase transaction is complete. To maximize the total revenue, the items presented on the recommendation page need to be thoughtfully chosen and placed. Meanwhile, the recommender system will learn the preference of the customers by observing their behaviors. In this scenario, the actions shown in Fig. 11 are the items for the recommender systems, and the states are the customer preferences learned by the system. In the following section, we will introduce an instance of reinforcement learning paradigm known as multi-armed bandit. We will discuss issues that may arise in  of multi-armed bandit models such as exploration and exploitation dilemma and confounding issue. We finally describe how Thompson sampling and causal Thompson sampling would help resolve such shortcomings. 

\subsubsection{Multi-armed bandit model}

The multi-armed bandit model is inspired by imagining a gambler sitting in front of a row of slot machines, and deciding which machine to play with the goal of maximizing his/her total return. Let us say we have $N$ slot machines. Each slot machine gives a positive reward with probability $p$, or a negative reward with probability $1-p$. When a gambler just starts his game, he has no knowledge of those slot machines. As he plays a few hands, he would “learn” which slot machine is more rewarding (i.e., the reward distribution of these actions). For example, if there are 5 slot machines with the reward distribution of $[0.1, 0.2, 0.3, 0.4, 0.5]$, the gambler would always choose the last machine to play in order to have the maximum return. However, the gambler does not have this reward information at the beginning of the game unless he or she pulls and tries each slot machine and estimate the reward.

Multi-armed bandit models have been widely adopted in the recommender systems. For example, an online video recommender system (e.g., YouTube) would pick a set of videos based on user watch history and expect high click-through-rate from the user side. In this scenario, the recommender system is the gambler and the videos are slot machines. The reward of each video is the user click. 

\begin{figure}[ht]
    \centering
    \includegraphics[width=70mm]{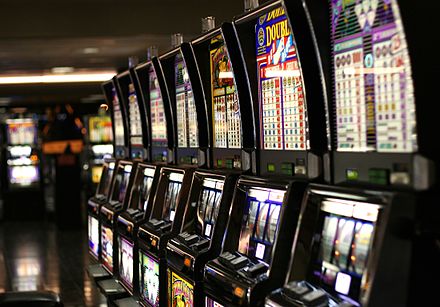}
    \caption{A row of slot machines in Las Vegas.}
    \label{fig:confounder}
\end{figure}{}

\subsubsection{Exploration and exploitation dilemma}
\label{sec:exploration_exploitation}

Before we have the complete reward distribution of our actions, we would face the problem of which action to take based on our partial observation. If we take a greedy approach of always choosing the action with the maximum reward we observed, this strategy might not be optimal. Let us understand it with the previous slot machine example where the true reward distribution is $[0.1, 0.2, 0.3, 0.4, 0.5]$. Imagine the gambler just sits in and only pulls a few arms, his reward estimation of these machines would be: $[0.5, 0.0, 0.0, 0.0, 0.0]$. This information may convince him to choose the first arm to play even though the likelihood of reward of this machine is lowest among all. The rest of the machines would never have the chance to be explored as their reward estimation are 0s. Therefore, we need to be careful of getting stuck in exploiting a specific action and forget to explore the unknowns. Over the years, there has been many proposals to handle this dilemma: 1) $\epsilon$-greedy algorithm, 2) upper confidence bound algorithm, 3) Thompson sampling algorithm, and many other variants. We will get back to this dilemma shortly. 

\subsection{Confounding issue in reinforcement learning}

How to view the reinforcement learning with the causal lens? Recall that all actions in reinforcement learning are associated with certain reward. If we see the actions as treatments, we can view the reward as a causal effect of the outcome from each action taken.

\begin{figure}[ht]
    \centering
    \includegraphics[width=70mm]{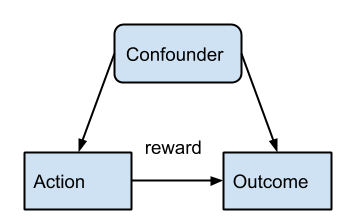}
    \caption{A graphical illustration demonstrates confounding bias in reinforcement learning.}
    \label{fig:confounding_rl}
\end{figure}

In section \ref{chap:chapter_deconfound}, we learned that confounders may bias the causal effect estimation given scenarios depicted in Figure \ref{fig:confounding_rl}. The biased reward estimation would lead to a suboptimal policy. Therefore, it is essential to debias the reward estimation in the existence of confounders. \cite{bareinboim2015bandits} proposes a causal Thompson Sampling technique which draws its strength upon “counterfactual reasoning”. Specifically, causal Thompson Sampling algorithm enables reinforcement learning agents to compare the rewards resulting from the action that the agent is about to take and the alternative actions that are overlooked.

{}

\subsubsection{Thompson sampling}

In multi-armed bandit problems \cite{russo2017tutorial}, the goal of maximizing the reward can be fulfilled in different ways. For instance, in our slot machine example, the gambler wants to estimate the true reward for each slot machine to make optimal choices and to maximum his profit in the end. One way is for the gambler to estimate and update the probability of winning at each machine based on his plays. The gambler would choose the slot machine with the maximum likelihood of winning in the next round. 
In Thompson sampling algorithm, however, maximizing reward, unlike the greedy way, is done by balancing the exploitation and exploration. The Thompson algorithm associates each machine with a beta probability distribution, which is essentially the distribution of success versus failure in slot machine events. In each turn, the algorithm will sample from associated beta distributions to generate the reward estimations, and then chooses to play the machine with the highest estimated reward. Beta distribution is parameterized by $\alpha$ and $\beta$ in the following equation,

\begin{equation}
    B(\alpha, \beta) = \frac{\Gamma(\alpha)\Gamma(\beta)}{\Gamma(\alpha+\beta)}
\end{equation}

where the $\Gamma$ is the Gamma distribution.  Let us inspect the property of beta distribution with the following examples and understand why it is a good candidate for the reward estimation in our slot machine example. In Figure 14 (c), we can see that if we sample from the beta distribution when $\alpha = \beta = 5$, the estimated reward would most likely be around 0.5. When $\alpha = 5$ and $\beta = 2$, the peak moves towards the right and the sample values might be around 0.8 in most cases (See Figure 14 (a)). If we treat $\alpha$ and $\beta$ as the number of successes and failures respectively in our casino game, then the larger value of $\alpha$ in a beta distribution indicates higher likelihood of payout, whereas the value of $\beta$ in a beta distribution represents lower likelihood of payout.

\begin{figure}
    \centering
    \includegraphics[width=100mm]{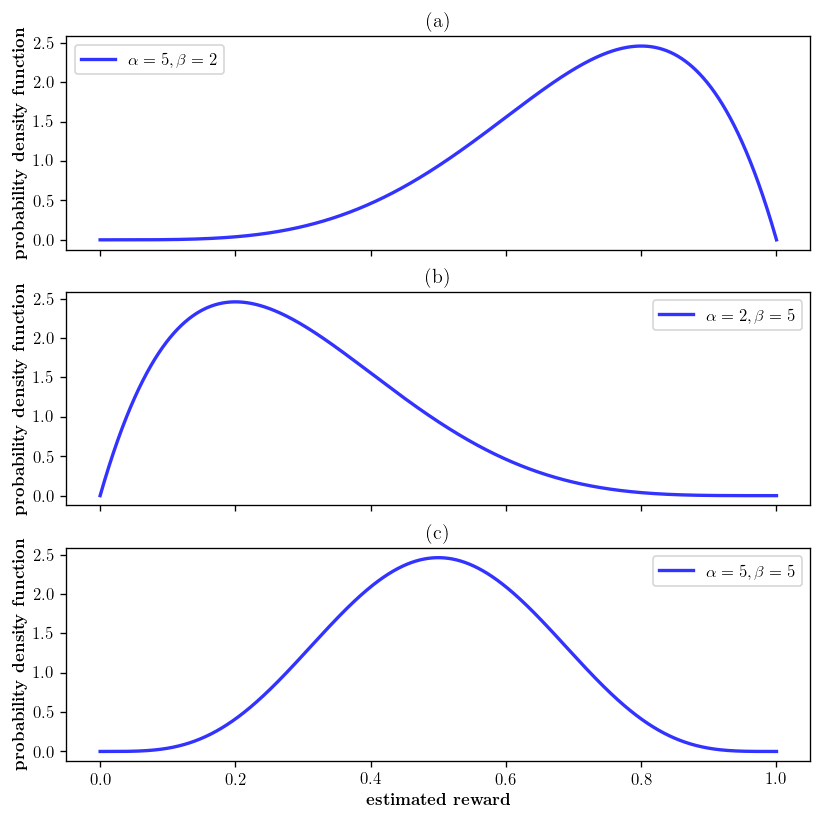}
    \caption{Beta distribution with varying $\alpha$ and $\beta$}
    \label{fig:beta_dist}
\end{figure}

In the Thompson Sampling algorithm, these beta distributions would be updated by maintaining the current number of successes ($\alpha$) and failures ($\beta$). As we can see from Figure 14 (b), we can still sample a large reward with small likelihood. This gives the opportunity for the unrewarding machines to be chosen. Therefore, Thompson Sampling algorithm addresses the exploration and exploitation dilemma in certain degree  \cite{russo2017tutorial}.

\subsubsection{Causal Thompson sampling} 

In \cite{bareinboim2015bandits}, Bareinboim et al. argues that the reward estimation in typical reinforcement learning may be biased in the existence of unobserved confounders. Therefore, conventional Thompson Sampling may choose these machines that are sub-optimal and skip the real optimal choices. The authors proposed an augmented Thompson Sampling with counterfactual reasoning approach, which allows us to reason about a scenario that has not happened. Imagine that a gambler sits in front of two slot machines (M1 has higher payout than M2). Say that the gambler is less confident in the reward estimation for the first machine than the second one. His natural predilection would be choosing the second slot machine (M2) as the gambler “believes” that the expected reward from the second machine is higher. Recall that unobserved confounders exist and may mislead the gambler believing that the second machine is more rewarding, whereas the first machine is actually the optimal choice. Therefore, if the gambler can answer these two questions: “given that I believe M2 is better, what the payout would be if I played M2?” (intuition), and “given that I believe M2 is better, what the payout would be if I acted differently?” (counter-intuition). If the gambler can estimate the rewards of the intuition and counter-intuition, he may rectify his misbelieve in the existence of confounders and choose the optimal machine. Interested readers can refer to \cite{bareinboim2015bandits} for more details. Experimental results in \cite{bareinboim2015bandits} show that Causal Thompson sampling solves the confounding issue with a toy example dataset. However, this work has not been validated on a realistic dataset. It is still not clear how to handle the confounding issue in real practice. Therefore, this is still an open research question.

\section{How to discover the causal relations from data?}
In the previous sections, we have seen a number of examples where the causal diagrams are essential in the causal reasoning. For example, we use the causal graphs in section 3 to demystify the confounding bias (e.g., Simpson Paradox). The causal graph in Fig. 4 was constructed after careful inspection in the data set with the help of domain experts. Some readers may wonder if the constructions of these causal structures can be automated with minimum inputs and interventions from experts. In this section, we introduce the concept of causal discovery which identifies these causal relations from the observational dataset via some computational algorithms \cite{glymour2019review}. 

\begin{figure}[ht]
    \centering
    \includegraphics[width=70mm]{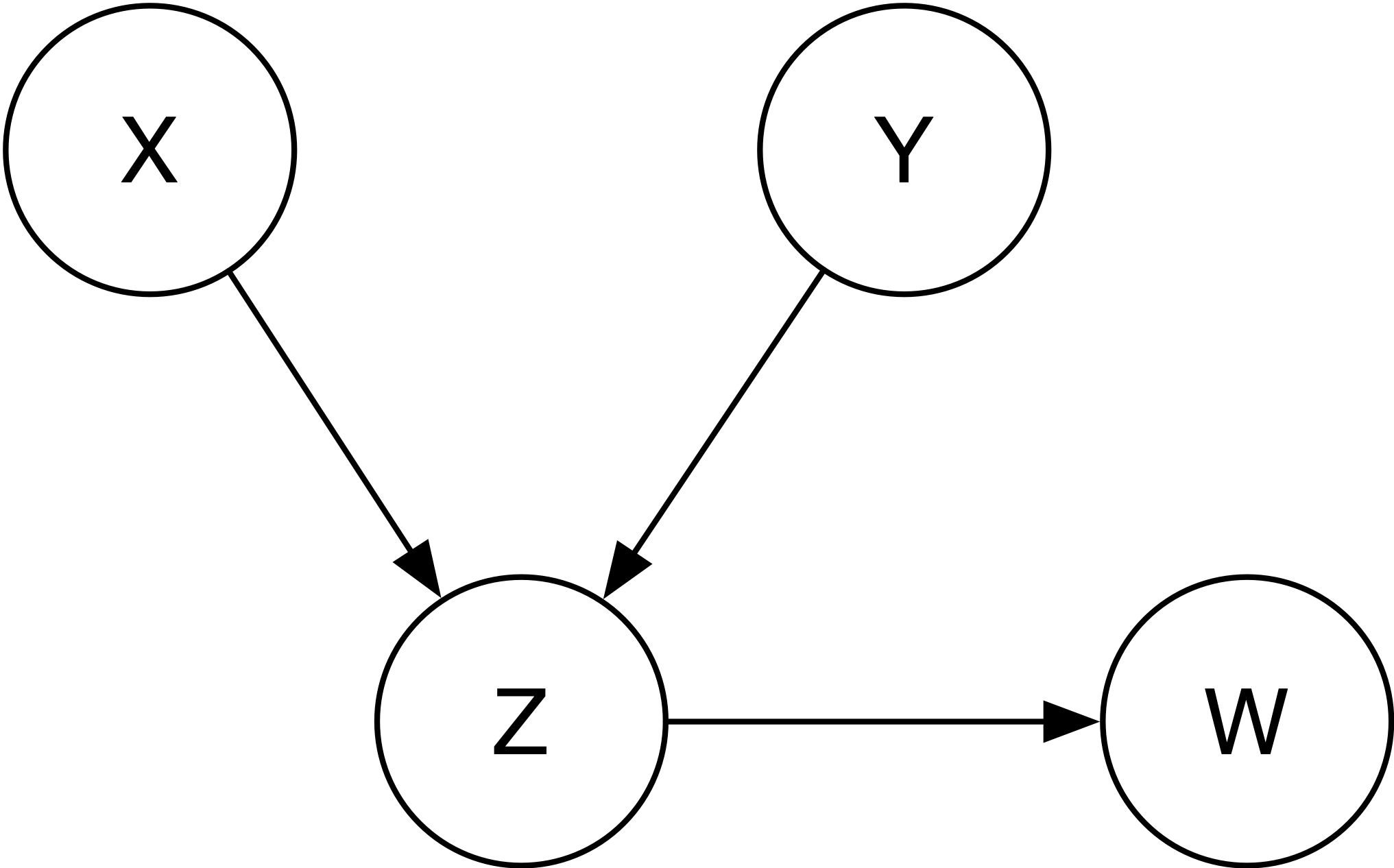}
    \caption{A causal graph used for causal discovery illustration.}
    \label{fig:causal_discovery}
\end{figure}{}

To recapitulate, a causal diagram encodes our assumptions about the causal story behind a dataset. These causal assumptions govern the data generation process. Therefore, the dataset should exhibit certain statistical properties that agree with the causal diagram. In particular, if we believe that a causal diagram $G$ might have generated a dataset $D$, the variables that are independent from each other in $G$ should also be tested as independent in the dataset $D$. For example, the causal diagram in Fig. 16 indicates that $X$ and $W$ are conditionally independent given $Z$ as “$X\ra Z \ra W$” forms a chain structure (refer to the d-separation in section 2.3.2). If a dataset is generated by the causal diagram in Fig. 16, the data should suggest that $X$ and $W$ are independent conditional on $Z$, written as $X \perp \!\!\! \perp W |Z$. On the contrary, we can refute this causal diagram when the data suggest that this conditional independence (e.g., $X \perp \!\!\! \perp W |Z$) does not hold. Granted that the dataset $D$ matches with every conditional independence relations suggested in the causal diagram $G$. Then we say the causal model $G$ is plausible for the dataset $D$. The task of causal discovery is to search through all the possible causal graphs and find the most plausible candidate. 

\begin{figure}[ht]
    \centering
    \includegraphics[width=120mm]{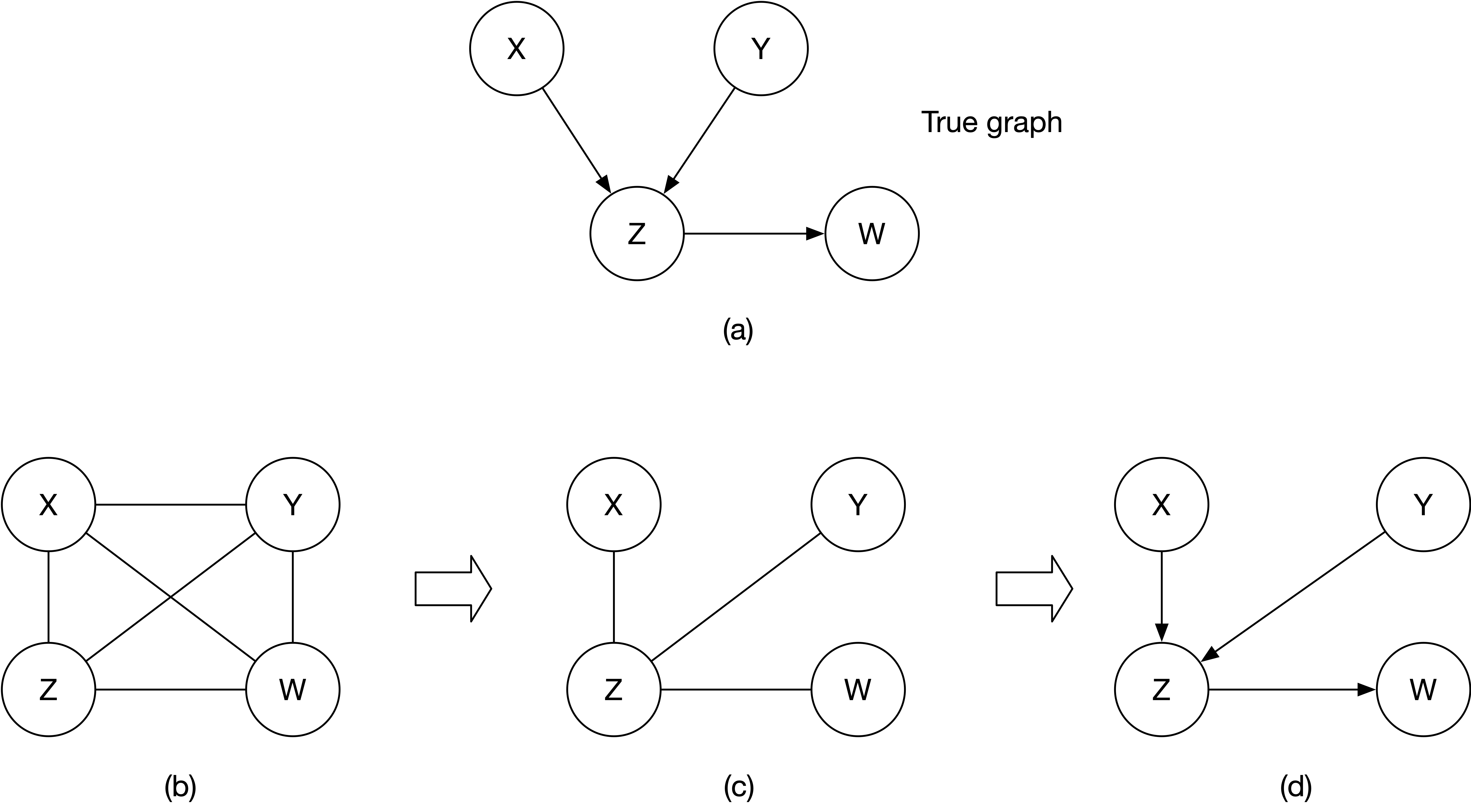}
    \caption{A simple example that illustrates PC algorithm.  (a) is the true causal structure. (b) PC algorithm starts with a fully connected graph. (c) It removes the edges if the data suggest that the linked variables are independent, e.g., X and Y are independent. (d) It decides the causal directions using d-separation..}
    \label{fig:PC_algorithm}
\end{figure}{}

The mainstream of causal discovery methods can be categorized into two groups, constraint-based methods or score-based methods. Constraint-based approaches exploit the conditional independence test to search the causal structure. Typical algorithms include PC algorithm \cite{spirtes2000causation} and FCI algorithm \cite{spirtes2000constructing}.  PC algorithm (named after its authors, Peter and Clark) starts the search with a fully connected graph as shown in Figure 17 (b). It repeatedly eliminates edges between two variables if they are independent in the data as shown in Fig. 17 (c). For example, the statistical testing checks whether $Prob(X,Y)=Prob(X) \times Prob(Y)$ holds. If the equation is valid in the data, then we conclude X and Y are independent. Finally, the PC algorithm decides the arrow directions using d-separations (Fig. 17 (d)). Particularly, the algorithm finds the v-structure “$X\ra Z \la Y$” when the data suggest that $X$ and $Y$ become dependent if conditioning on $Z$. This procedure is done in two steps: 1) verify that $Prob(X,Y)=Prob(X) \times Prob(Y)$; 2) verify that $Prob(X,Y | Z) \neq Prob(X | Z) \times Prob(Y | Z)$. The first step indicates that X and Y are initially independent if not conditioning on $Z$. The second step tells that $X$ and $Y$ become dependent conditional on $Z$. Moreover, PC algorithm identifies the chain structure “$X \ra Z \ra W$” when the data suggest that $X$ and $W$ are independent conditional on $Z$, that is $Prob(X,W | Z)=Prob(X | Z) \times Prob(W | Z)$. The chain structure “$Y \ra Z \ra W$” is also identified in the similar way. Score-based approach for causal discovery aims to optimize a predefined score function (e.g., Bayesian information criteria) that evaluates the fitness of a candidate causal graph against the given dataset.   A common score-based method is fast greedy equivalence search (FGES) \cite{ramsey2017million}. This method starts the search with an empty graph and iteratively adds an directed edge in the existing structure if the fitness of the current graph is increased. This process continues till the fitness score is no longer improved. Next, the algorithm begins removing an edge if the deletion improves the score. The edge removal stops when the score cannot be improved further. Then the resulting graph is a plausible causal structure from the dataset \cite{chickering2002optimal}.

However, causal structure learning with high dimensional dataset has been deemed challenging since the search space of directed acyclic graphs scales super-exponentially with the number of nodes. Generally, the search problem of identifying DAGs from data has been considered as NP-hard \cite{chickering1996learning,chickering2004large}. This challenge greatly handicaps the wide usage of the aforementioned causal discovery methods. There exists ongoing efforts to make the search more efficient by relaxing certain assumptions \cite{claassen2013learning}, adopting parallelized computation methodology \cite{le2016fast}, and incorporating deep learning framework \cite{ng2019graph}. The detailed discussion on causal discovery is beyond the scope of this chapter. Those interested in learning more about causal discovery should refer to \cite{ramsey2017million,chickering2002optimal,chickering1996learning,chickering2004large,claassen2013learning,le2016fast,ng2019graph}.

\section{Conclusions}

In this chapter, we explored the strengths of causal reasoning when facing problems such as confounding bias, model transportability, and learning from missing data. We present some examples to demonstrate pure data-driven or correlation-based statistical analysis may generate misleading conclusions. We argued the need to consider causality in our models to support critical clinical decision-making. Machine learning has been widely employed in various healthcare applications with recently increased efforts on how to augment machine learning models with causality to improve interpretability \cite{kim2017interpretable,moraffah2020causal} and predictive fairness \cite{zhang2020large,schnabel2016recommendations} and to avoid bias \cite{kusner2017counterfactual,loftus2018causal}. The model interpretability can be enhanced through the identification of cause-effect relation between the model input and outcome. We can observe how the model outcome responds to interventions upon inputs. For example, powerful machine learning models can be built for early detection of type 2 diabetes mellitus using a collection of features such as age, weight, HDL cholesterol, and triglycerides \cite{kopitar2020early}. However, healthcare practitioners are not content with mere predictions – they are also interested in the variables upon which the intervention will help reduce the risk of the disease effectively. Understanding causality is crucial to answer such questions. We also showed how causality can address confounding bias and selection bias in data analyses. Literature shows that causal inference can be adopted in deep learning modeling to reduce selection bias in recommender systems \cite{zhang2020large,schnabel2016recommendations}. Model fairness aims to protect the benefit of people in the minority groups or historically disadvantageous groups from the discriminative decisions produced by AI. Causal inference can also ensure model fairness against such social discriminations \cite{kusner2017counterfactual}. In addition to the attempts and progressed made in this field, there are many low-hanging fruits in combining causal inference with machine learning methods. We hope this brief introduction of causal inference can inspire more interested readers in this research area.

\bibliographystyle{unsrt}
\bibliography{references}

\begin{thebibliography}{10}

\bibitem{geer2011correlation}
Daniel~E Geer~Jr.
\newblock Correlation is not causation.
\newblock {\em IEEE Security \& Privacy}, 9(2):93--94, 2011.

\bibitem{havens1999correlation}
KE~Havens.
\newblock Correlation is not causation: a case study of fisheries, trophic
  state and acidity in florida (usa) lakes.
\newblock {\em Environmental Pollution}, 106(1):1--4, 1999.

\bibitem{hume2016enquiry}
David Hume.
\newblock An enquiry concerning human understanding.
\newblock In {\em Seven masterpieces of philosophy}, pages 191--284. Routledge,
  2016.

\bibitem{pearl2018book}
Judea Pearl and Dana Mackenzie.
\newblock {\em The book of why: the new science of cause and effect}, chapter
  Beyond Adjustment: The Conquest of Mount Intervention, page 234.
\newblock Basic Books, 2018.

\bibitem{morabia2005epidemiological}
Alfredo Morabia.
\newblock Epidemiological causality.
\newblock {\em History and philosophy of the life sciences}, pages 365--379,
  2005.

\bibitem{parascandola2011causes}
Mark Parascandola.
\newblock Causes, risks, and probabilities: probabilistic concepts of causation
  in chronic disease epidemiology.
\newblock {\em Preventive Medicine}, 53(4-5):232--234, 2011.

\bibitem{chocolate}
Franz~H. Messerli.
\newblock Chocolate consumption, cognitive function, and nobel laureates.
\newblock {\em New England Journal of Medicine}, 367(16):1562--1564, 2012.
\newblock PMID: 23050509.

\bibitem{vanderweele2013defi}
Tyler~J VanderWeele and Ilya Shpitser.
\newblock On the definition of a confounder.
\newblock {\em Annals of statistics}, 41(1):196, 2013.

\bibitem{julious1994confounding}
Steven~A Julious and Mark~A Mullee.
\newblock Confounding and simpson's paradox.
\newblock {\em Bmj}, 309(6967):1480--1481, 1994.

\bibitem{holland1983lord}
Paul~W Holland and Donald~B Rubin.
\newblock On lord's paradox.
\newblock {\em Principals of modern psychological measurement}, pages 3--25,
  1983.

\bibitem{greenland2001confounding}
Sander Greenland and Hal Morgenstern.
\newblock Confounding in health research.
\newblock {\em Annual review of public health}, 22(1):189--212, 2001.

\bibitem{pearl2009causal}
Judea Pearl et~al.
\newblock Causal inference in statistics: An overview.
\newblock {\em Statistics surveys}, 3:96--146, 2009.

\bibitem{hunermund2019causal}
Paul H{\"u}nermund and Elias Bareinboim.
\newblock Causal inference and data-fusion in econometrics.
\newblock {\em arXiv preprint arXiv:1912.09104}, 2019.

\bibitem{verde2011smoking}
Zoraida Verde, Catalina Santiago, Jos{\'e} Miguel~Rodr{\'\i}guez
  Gonz{\'a}lez-Moro, Pilar de~Lucas~Ramos, Soledad~L{\'o}pez Mart{\'\i}n,
  Fernando Bandr{\'e}s, Alejandro Lucia, and F{\'e}lix G{\'o}mez-Gallego.
\newblock ‘smoking genes’: a genetic association study.
\newblock {\em PloS one}, 6(10):e26668, 2011.

\bibitem{mackillop2010role}
James MacKillop, Ezemenari~M Obasi, Michael~T Amlung, John~E McGeary, and
  Valerie~S Knopik.
\newblock The role of genetics in nicotine dependence: mapping the pathways
  from genome to syndrome.
\newblock {\em Current cardiovascular risk reports}, 4(6):446--453, 2010.

\bibitem{pearl2012calculus}
Judea Pearl.
\newblock The do-calculus revisited.
\newblock {\em arXiv preprint arXiv:1210.4852}, 2012.

\bibitem{tucci2013introduction}
Robert~R Tucci.
\newblock Introduction to judea pearl's do-calculus.
\newblock {\em arXiv preprint arXiv:1305.5506}, 2013.

\bibitem{huang2012pearl}
Yimin Huang and Marco Valtorta.
\newblock Pearl's calculus of intervention is complete.
\newblock {\em arXiv preprint arXiv:1206.6831}, 2012.

\bibitem{geiger1990identifying}
Dan Geiger, Thomas Verma, and Judea Pearl.
\newblock Identifying independence in bayesian networks.
\newblock {\em Networks}, 20(5):507--534, 1990.

\bibitem{pearl1985bayesian}
Judea Pearl.
\newblock Bayesian networks: A model of self-activated memory for evidential
  reasoning.
\newblock In {\em Proceedings of the 7th Conference of the Cognitive Science
  Society, University of California, Irvine, CA, USA}, pages 15--17, 1985.

\bibitem{pearl2001bayesian}
Judea Pearl.
\newblock Bayesian networks, causal inference and knowledge discovery.
\newblock {\em UCLA Cognitive Systems Laboratory, Technical Report}, 2001.

\bibitem{heckerman1999bayesian}
David Heckerman, Christopher Meek, and Gregory Cooper.
\newblock A bayesian approach to causal discovery.
\newblock {\em Computation, causation, and discovery}, 19:141--166, 1999.

\bibitem{vanderweele2013definition}
Tyler~J VanderWeele and Ilya Shpitser.
\newblock On the definition of a confounder.
\newblock {\em Annals of statistics}, 41(1):196, 2013.

\bibitem{greenland1999confounding}
Sander Greenland, James~M Robins, and Judea Pearl.
\newblock Confounding and collapsibility in causal inference.
\newblock {\em Statistical science}, pages 29--46, 1999.

\bibitem{charig1986comparison}
Clive~R Charig, David~R Webb, Stephen~Richard Payne, and John~E Wickham.
\newblock Comparison of treatment of renal calculi by open surgery,
  percutaneous nephrolithotomy, and extracorporeal shockwave lithotripsy.
\newblock {\em Br Med J (Clin Res Ed)}, 292(6524):879--882, 1986.

\bibitem{pearl2016causal}
Judea Pearl, Madelyn Glymour, and Nicholas~P Jewell.
\newblock {\em Causal inference in statistics: A primer}, chapter The effects
  of Interventions, pages 53--55.
\newblock John Wiley \& Sons, 2016.

\bibitem{murphy2012machine}
Kevin~P Murphy.
\newblock {\em Machine learning: a probabilistic perspective}.
\newblock MIT press, 2012.

\bibitem{EBtalkColumbia}
{Elias Bareinboim -- Causal Data Science}.
\newblock \url{https://www.youtube.com/watch?v=dUsokjG4DHc}, 2019.

\bibitem{bareinboim2016causal}
Elias Bareinboim and Judea Pearl.
\newblock Causal inference and the data-fusion problem.
\newblock {\em Proceedings of the National Academy of Sciences},
  113(27):7345--7352, 2016.

\bibitem{bareinboim2012controlling}
Elias Bareinboim and Judea Pearl.
\newblock Controlling selection bias in causal inference.
\newblock In {\em Artificial Intelligence and Statistics}, pages 100--108,
  2012.

\bibitem{NYC}
Jeremy Berke.
\newblock A statewide antibody study estimates that 21\% of new york city
  residents have had the coronavirus, cuomo says, 2020.

\bibitem{NYC_selection_bias}
Cassie Kozyrkov.
\newblock Were 21\% of new york city residents really infected with the novel
  coronavirus?, 2020.

\bibitem{zhang2019causal}
Wenhao Zhang, Wentian Bao, Xiao-Yang Liu, Keping Yang, Quan Lin, Hong Wen, and
  Ramin Ramezani.
\newblock A causal perspective to unbiased conversion rate estimation on data
  missing not at random, 2019.

\bibitem{ryder2011advantage}
Anthony~B Ryder, Anna~V Wilkinson, Michelle~K McHugh, Katherine Saunders,
  Sumesh Kachroo, Anthony D’Amelio, Melissa Bondy, and Carol~J Etzel.
\newblock The advantage of imputation of missing income data to evaluate the
  association between income and self-reported health status (srh) in a mexican
  american cohort study.
\newblock {\em Journal of immigrant and minority health}, 13(6):1099--1109,
  2011.

\bibitem{rubin1976inference}
Donald~B Rubin.
\newblock Inference and missing data.
\newblock {\em Biometrika}, 63(3):581--592, 1976.

\bibitem{enders2010applied}
Craig~K Enders.
\newblock {\em Applied missing data analysis}.
\newblock Guilford press, 2010.

\bibitem{mohan2013graphical}
Karthika Mohan, Judea Pearl, and Jin Tian.
\newblock Graphical models for inference with missing data.
\newblock In {\em Advances in neural information processing systems}, pages
  1277--1285, 2013.

\bibitem{pearl2013recoverability}
Judea Pearl and Karthika Mohan.
\newblock Recoverability and testability of missing data: Introduction and
  summary of results.
\newblock {\em Available at SSRN 2343873}, 2013.

\bibitem{mohan2015missing}
Karthika Mohan and Judea Pearl.
\newblock Missing data from a causal perspective.
\newblock In {\em Workshop on Advanced Methodologies for Bayesian Networks},
  pages 184--195. Springer, 2015.

\bibitem{yoon2018gain}
Jinsung Yoon, James Jordon, and Mihaela Van Der~Schaar.
\newblock Gain: Missing data imputation using generative adversarial nets.
\newblock {\em arXiv preprint arXiv:1806.02920}, 2018.

\bibitem{buuren2010mice}
S~van Buuren and Karin Groothuis-Oudshoorn.
\newblock mice: Multivariate imputation by chained equations in r.
\newblock {\em Journal of statistical software}, pages 1--68, 2010.

\bibitem{deng2016multiple}
Yi~Deng, Changgee Chang, Moges~Seyoum Ido, and Qi~Long.
\newblock Multiple imputation for general missing data patterns in the presence
  of high-dimensional data.
\newblock {\em Scientific reports}, 6:21689, 2016.

\bibitem{schafer2002missing}
Joseph~L Schafer and John~W Graham.
\newblock Missing data: our view of the state of the art.
\newblock {\em Psychological methods}, 7(2):147, 2002.

\bibitem{little2019statistical}
Roderick~JA Little and Donald~B Rubin.
\newblock {\em Statistical analysis with missing data}, volume 793.
\newblock John Wiley \& Sons, 2019.

\bibitem{pearl1995testability}
Judea Pearl.
\newblock On the testability of causal models with latent and instrumental
  variables.
\newblock In {\em Proceedings of the Eleventh conference on Uncertainty in
  artificial intelligence}, pages 435--443. Morgan Kaufmann Publishers Inc.,
  1995.

\bibitem{ford2018architects}
Martin Ford.
\newblock {\em Architects of Intelligence: The truth about AI from the people
  building it}.
\newblock Packt Publishing Ltd, 2018.

\bibitem{bareinboim2015bandits}
Elias Bareinboim, Andrew Forney, and Judea Pearl.
\newblock Bandits with unobserved confounders: A causal approach.
\newblock In {\em Advances in Neural Information Processing Systems}, pages
  1342--1350, 2015.

\bibitem{russo2017tutorial}
Daniel Russo, Benjamin Van~Roy, Abbas Kazerouni, Ian Osband, and Zheng Wen.
\newblock A tutorial on thompson sampling.
\newblock {\em arXiv preprint arXiv:1707.02038}, 2017.

\bibitem{glymour2019review}
Clark Glymour, Kun Zhang, and Peter Spirtes.
\newblock Review of causal discovery methods based on graphical models.
\newblock {\em Frontiers in genetics}, 10:524, 2019.

\bibitem{spirtes2000causation}
Peter Spirtes, Clark~N Glymour, Richard Scheines, and David Heckerman.
\newblock {\em Causation, prediction, and search}.
\newblock MIT press, 2000.

\bibitem{spirtes2000constructing}
Pater Spirtes, Clark Glymour, Richard Scheines, Stuart Kauffman, Valerio
  Aimale, and Frank Wimberly.
\newblock Constructing bayesian network models of gene expression networks from
  microarray data.
\newblock 2000.

\bibitem{ramsey2017million}
Joseph Ramsey, Madelyn Glymour, Ruben Sanchez-Romero, and Clark Glymour.
\newblock A million variables and more: the fast greedy equivalence search
  algorithm for learning high-dimensional graphical causal models, with an
  application to functional magnetic resonance images.
\newblock {\em International journal of data science and analytics},
  3(2):121--129, 2017.

\bibitem{chickering2002optimal}
David~Maxwell Chickering.
\newblock Optimal structure identification with greedy search.
\newblock {\em Journal of machine learning research}, 3(Nov):507--554, 2002.

\bibitem{chickering1996learning}
David~Maxwell Chickering.
\newblock Learning bayesian networks is np-complete.
\newblock In {\em Learning from data}, pages 121--130. Springer, 1996.

\bibitem{chickering2004large}
Max Chickering, David Heckerman, and Chris Meek.
\newblock Large-sample learning of bayesian networks is np-hard.
\newblock {\em Journal of Machine Learning Research}, 5, 2004.

\bibitem{claassen2013learning}
Tom Claassen, Joris Mooij, and Tom Heskes.
\newblock Learning sparse causal models is not np-hard.
\newblock {\em arXiv preprint arXiv:1309.6824}, 2013.

\bibitem{le2016fast}
Thuc~Duy Le, Tao Hoang, Jiuyong Li, Lin Liu, Huawen Liu, and Shu Hu.
\newblock A fast pc algorithm for high dimensional causal discovery with
  multi-core pcs.
\newblock {\em IEEE/ACM transactions on computational biology and
  bioinformatics}, 16(5):1483--1495, 2016.

\bibitem{ng2019graph}
Ignavier Ng, Shengyu Zhu, Zhitang Chen, and Zhuangyan Fang.
\newblock A graph autoencoder approach to causal structure learning.
\newblock {\em arXiv preprint arXiv:1911.07420}, 2019.

\bibitem{kim2017interpretable}
Jinkyu Kim and John Canny.
\newblock Interpretable learning for self-driving cars by visualizing causal
  attention.
\newblock In {\em Proceedings of the IEEE international conference on computer
  vision}, pages 2942--2950, 2017.

\bibitem{moraffah2020causal}
Raha Moraffah, Mansooreh Karami, Ruocheng Guo, Adrienne Raglin, and Huan Liu.
\newblock Causal interpretability for machine learning-problems, methods and
  evaluation.
\newblock {\em ACM SIGKDD Explorations Newsletter}, 22(1):18--33, 2020.

\bibitem{zhang2020large}
Wenhao Zhang, Wentian Bao, Xiao-Yang Liu, Keping Yang, Quan Lin, Hong Wen, and
  Ramin Ramezani.
\newblock Large-scale causal approaches to debiasing post-click conversion rate
  estimation with multi-task learning.
\newblock In {\em Proceedings of The Web Conference 2020}, pages 2775--2781,
  2020.

\bibitem{schnabel2016recommendations}
Tobias Schnabel, Adith Swaminathan, Ashudeep Singh, Navin Chandak, and Thorsten
  Joachims.
\newblock Recommendations as treatments: Debiasing learning and evaluation.
\newblock In {\em international conference on machine learning}, pages
  1670--1679. PMLR, 2016.

\bibitem{kusner2017counterfactual}
Matt~J Kusner, Joshua~R Loftus, Chris Russell, and Ricardo Silva.
\newblock Counterfactual fairness.
\newblock {\em arXiv preprint arXiv:1703.06856}, 2017.

\bibitem{loftus2018causal}
Joshua~R Loftus, Chris Russell, Matt~J Kusner, and Ricardo Silva.
\newblock Causal reasoning for algorithmic fairness.
\newblock {\em arXiv preprint arXiv:1805.05859}, 2018.

\bibitem{kopitar2020early}
Leon Kopitar, Primoz Kocbek, Leona Cilar, Aziz Sheikh, and Gregor Stiglic.
\newblock Early detection of type 2 diabetes mellitus using machine
  learning-based prediction models.
\newblock {\em Scientific reports}, 10(1):1--12, 2020.

\end{thebibliography}

\end{document}